\documentclass{bmvc2k}

\title{AlloEgo-VLM: Disambiguating Allocentric and Egocentric Reference Frames in Vision--Language Models}

\addauthor{ Kuan-Lin Chen (Student)}{andy90123.ai12@nycu.edu.tw}{1}
\addauthor{ Tzu-Ti Wei (Student)}{a2699560.ai09@nycu.edu.tw}{1}
\addauthor{ Chao-Chi Liao (Student)}{apricity.cs13@nycu.edu.tw}{1}
\addauthor{ Yu-Chee Tseng (Prof)}{https://sites.google.com/view/yctseng}{1}
\addauthor{ Jen-Jee Chen (Prof)}{https://people.cs.nycu.edu.tw/~chencz/}{1}

\addinstitution{\small
  \hspace*{-1cm} Institute of Computer Science and Engineering \\
  \hspace*{-1cm} \& College of Artificial Intelligence \\
  \hspace*{-1cm} National Yang Ming Chiao Tung University \\
  \hspace*{-1cm} Hsinchu, Taiwan \\[1em]
  \hspace*{-1cm} \textcolor{blue}{https://github.com/CKL9001/AlloEgo-VLM.git}
}

\runninghead{Chen et al.}{AlloEgo-VLM}

\usepackage{graphicx}
\usepackage{subcaption}
\usepackage{multirow}
\usepackage{booktabs}
\usepackage{xcolor}

\usepackage{enumitem}

\usepackage{algorithm}
\usepackage{algpseudocode}
\usepackage{amsmath}
\usepackage{graphics}
\usepackage{epsfig}
\usepackage{colortbl}
\usepackage{appendix}
\usepackage{hyperref}
\hypersetup{
    colorlinks=true,
    urlcolor=blue
}
\usepackage{wrapfig}
\usepackage{caption}
\usepackage{subcaption}
\usepackage{soul}

\definecolor{mycaptionblue}{RGB}{0, 0, 128}

\usepackage{amssymb}
\usepackage{graphics}
\usepackage{epsfig}
\usepackage{amsmath}
\usepackage{microtype}
\begin{document}
\maketitle

\begin{abstract}
This study investigates the challenge of ambiguity faced by Vision--Language Models (VLMs) in understanding spatial semantics.  
Spatial cognition, shaped by cognitive psychology, spatial science, and cultural context, often assigns directionality to objects.  
However, natural language descriptions of spatial relations frequently omit explicit reference frames, leading to semantic ambiguity and potentially serious errors for embodied AI robots.  
Existing VLMs, due to insufficient training on reference frames and object orientations, often produce inconsistent responses.  
To address this issue, we construct a new dataset, \emph{AlloEgo-View}, comprising (image, query, view-specific answer) triplets that capture key object relations from both allocentric and egocentric perspectives.
The view-specific descriptions follow a structured spatial representation that annotate detailed scene descriptions, reference and target objects, their orientations, reference frames, and view types. 
Building on AlloEgo-View, we develop \emph{AlloEgo-VLM}, a framework to disambiguate allocentric and egocentric reference frames, even under ambiguous queries, and to be easily integrated into existing VLMs via supervised fine-tuning.  
Furthermore, we deploy our framework onto an embodied robotic platform within NVIDIA Isaac Sim to validate its real-world feasibility in open-ended object searching tasks.
Experiments highlight the limitations of current VLMs in handling view-specific queries and demonstrate the strong disambiguation ability of AlloEgo-VLM.
\vspace{+60pt}
\end{abstract}

\section{Introduction}
\label{sec:intro}

\begin{wrapfigure}{r}{0.6\textwidth} 
    \centering
    \vspace{-10pt}
    \includegraphics[width=0.58\textwidth]{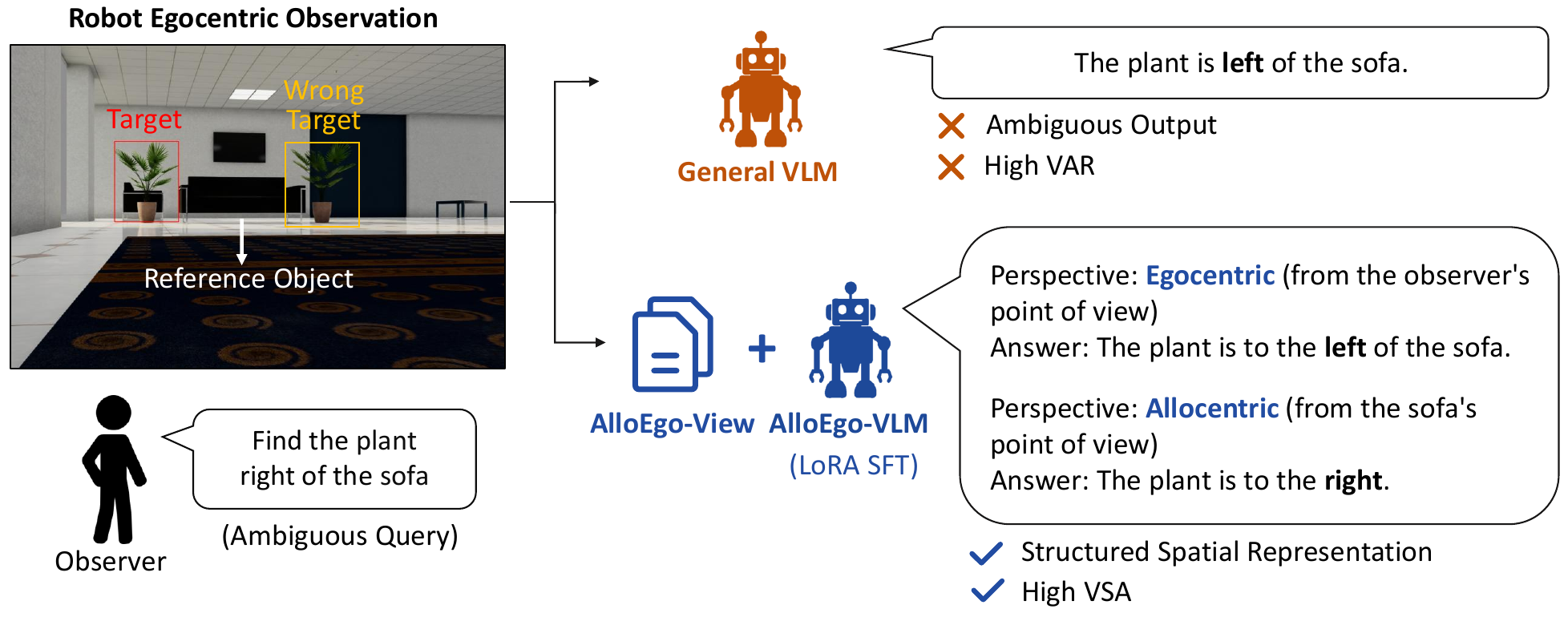}
    \vspace{9pt}
    \captionsetup{labelfont={color=mycaptionblue,bf}, textfont={color=mycaptionblue}}
    \caption{Under an ambiguous instruction, general VLMs suffer from high view ambiguity rate (VAR). AlloEgo-VLM resolves this by decoupling spatial semantics into explicit egocentric and allocentric perspectives, ensuring precise view-specific accuracy (VSA).}
    \label{fig:semantic_ambiguity}
    \vspace{-10pt}
\end{wrapfigure}

Understanding spatial semantics is a fundamental capability for Vision--Language Models (VLMs), particularly in human--robot interaction and embodied AI for navigation and manipulation tasks~\cite{Chen2024SpatialVLM, Ji2025CoTSpatial}.  
However, spatial descriptions in natural language are inherently ambiguous due to the diversity of reference frames employed by humans.  
Research in cognitive psychology and spatial science has shown that people often omit explicit reference frames in conversation, instead relying on context or cultural conventions~\cite{Janzen2012FrameNeural, Levinson2022Space, Levinson1996Absolute}.  
As illustrated in Fig.~\ref{fig:semantic_ambiguity}, reference object \textit{sofa} is facing the observer. \textit{``Find the plant right of the sofa''} yields distinct interpretations depending on whether an egocentric or allocentric perspective is assumed~\cite{Filimon2015Egocentric}. General VLMs regularly fail due to shallow image-space heuristics; they default to the observer's viewpoint, producing an uncalibrated response (\textit{``The plant is left of the sofa''}) that incorrectly anchors onto the \textit{Wrong Target}, leading to a high \textbf{View Ambiguity Rate (VAR)}. Conversely, AlloEgo-VLM suppresses this ego-bias through a dual-perspective \textbf{Structured Spatial Representation}. By decoupling the relations—identifying the object as egocentric left but allocentric right—our model correctly grounds the true \textit{Target} and achieves superior \textbf{View-Specific Accuracy (VSA)}.

Similar challenges have been noted in prior works~\cite{SPHERE2024} on robot navigation~\cite{ThinkingSpace2024, Li2025ViewSpatialBench}, {\em abstract perspective change}~\cite{PerspectiveAware2025}, and video action prediction~\cite{VLM4D2025}.  
These studies introduced datasets or benchmarks annotated with reference frames, providing valuable resources for spatial reasoning.  

However, their problem formulations often require users to explicitly specify the reference frame---either directly or implicitly---in their queries (e.g., \textit{``Find the plant right of the sofa from the perspective of the observer.''}). 
In other cases, the evaluations are framed as multiple-choice tasks.  
Such settings do not fully reflect real-world scenarios, where humans rarely phrase spatial questions in such explicit or verbose ways.

Our work does not seek to eliminate the inherent ambiguity that has existed in human language for centuries.  
Instead, it aims to enable robots and VLMs to handle naturally ambiguous queries without requiring users to explicitly specify the reference frame.  
Even for a simple question such as \textit{``Where is the plant?''}, the system should automatically resolve inconsistent interpretations and generate the correct answer across different reference frames.


Building on these observations, we note that existing VLMs and embodied AI systems are rarely trained with explicit annotations of reference frames or object orientations, often resulting in inconsistent or even contradictory spatial reasoning~\cite{Stogiannidis2025MindGap, Wang2024SpatialEval}.  
Quantitative analyses highlight the severity of this issue: in VSI-Bench (tiny), 71\% of errors stem from spatial reasoning, with nearly 22\% attributed to egocentric--allocentric transformations~\cite{ThinkingSpace2024}.  
In COMFORT++ and 3DSRBench, baseline VLMs perform close to random under allocentric views (LLaVA-NeXT 48\%, SpatialVLM 46\%, Molmo 36\%), reflecting ambiguity rates of 45--60\%~\cite{PerspectiveAware2025}.  
Our experiments further show that even state-of-the-art models collapse due to the lack of view awareness and disambiguation capabilities 
(refer to Sec.~\ref{sec:view-specific}).

Without awareness of the natural ambiguity in human language, an embodied robot may easily misinterpret spatial relations, leading to serious navigation errors or unsafe manipulations.  
The fundamental challenge of resolving orientation and direction ambiguity in real-world, unconstrained language remains largely unsolved.  
To address these issues, we construct a new dataset, \textbf{AlloEgo-View}, comprising (image, query, answer) triplets, where queries may contain natural linguistic ambiguity, while the answers are view-specific with explicit reference frames for disambiguating allocentric and egocentric viewpoints.  
Based on AlloEgo-View, we develop \textbf{AlloEgo-VLM}, a plug-and-play module with a \textbf{multi-stage training framework} that progressively expands view-specific data and fine-tunes VLMs to enhance spatial reasoning and view-specific grounding across perspectives, thereby improving the base model over successive stages.

Our main contributions are summarized as follows:
\begin{itemize}[noitemsep]
\item 
We identify and formalize the problem of \textbf{reference-frame ambiguity} in Vision--Language Models (VLMs) and highlight its implications for embodied AI.

\item 
We construct \textbf{AlloEgo-View}, a dataset with explicit allocentric and egocentric reference-frame annotations for spatial reasoning.

\item 
We propose \textbf{AlloEgo-VLM}, a plug-and-play framework with multi-stage training that progressively expands reference-frame-aware data and improves spatial grounding across perspectives.

\item 
Extensive experiments demonstrate improved spatial reasoning and reference-frame grounding over existing VLMs. Deployment in NVIDIA Isaac Sim further validates the framework in open-ended robotic object search tasks.
\end{itemize}


\section{Related Work}
\label{sec:related-work}

\subsection{Large Language and Vision--Language Models}

Large Language Models (LLMs) have achieved remarkable progress in tasks such as text summarization, question answering, and multi-step reasoning~\cite{Brown2020GPT3, meta2024llama3_2_v}.  
Instruction tuning has further aligned these models with human preferences~\cite{Wei2022FLAN, Ouyang2022InstructGPT}.  
Building on these advances, recent research has extended into multi-modality, giving rise to VLMs.  
Representative VLMs such as \textit{GPT-4o}~\cite{OpenAI2023GPT4}, \textit{LLaVA}~\cite{Liu2023LLaVA}, and \textit{InstructBLIP}~\cite{Dai2023InstructBLIP} demonstrate strong performance in various visual tasks.
Their ability to integrate linguistic reasoning with visual grounding makes them promising candidates for spatial relations reasoning.

\subsection{Spatial Understanding in VLMs}

Recent research has explored a wide spectrum of spatial understanding tasks that extend across both 2D and 3D domains. These include predicting object size and relative scale \cite{Yang2019SpatialSense}, estimating distances and depth relations \cite{Liao2024QSpatialBench}, and reasoning about directions or object localization within complex scenes \cite{Cheng2024SpatialRGPT}. In 3D settings, spatial reasoning further encompasses navigation, scene reconstruction, and embodied object interaction tasks \cite{SPHERE2024, tellex2011understanding}.

To support these tasks, a number of benchmarks and datasets have been developed. \textit{CLEVR} \cite{Johnson2017CLEVR} provides synthetic images with compositional object arrangements and detailed spatial relationships, enabling controlled reasoning tests. \textit{GQA} \cite{Hudson2019GQA} extends this to real-world images with structured scene graphs, supporting multi-step reasoning. \textit{SpatialSense} \cite{Yang2019SpatialSense} focuses on natural images with annotated spatial relationships, while \textit{SPAR} \cite{SPAR2025} captures relational ambiguity by providing multiple valid interpretations of spatial descriptions. \textit{ViewSpatial-Bench} \cite{Li2025ViewSpatialBench} and \textit{Spatial-Comfort} \cite{Liu2024SpatialFoRAmbiguities} tackle reasoning consistency under viewpoint variation.

Despite these advances, most existing approaches still treat spatial reasoning primarily as a geometric or relational problem, without addressing how linguistic descriptions vary across perspectives and contexts, such as allocentric and egocentric views.

\begin{table}[t]
\centering
\scriptsize
\resizebox{\linewidth}{!}{
    \begin{tabular}{|c|c|c|c|c|c|}
        \hline
        \textbf{Dataset} & \textbf {Task Description} & \textbf{View Type} & \textbf{MCAF} & \textbf{RFF} & \textbf{Scale} \\
        \hline
        CLEVR \cite{Johnson2017CLEVR} & Visual reasoning (synthetic QA) & Ego                      & $\triangle$ & $\checkmark$ & $\sim$700K \\
        \hline
        GQA \cite{Hudson2019GQA} & VQA + scene graph reasoning & Ego                               & $\triangle$ & $\checkmark$ & $\sim$22M \\
        \hline
        SpatialSense \cite{Yang2019SpatialSense} & Pairwise spatial relation classification & Ego  & $\checkmark$ & $\checkmark$ & $\sim$11K \\
        \hline
        SPAR \cite{SPAR2025} & Spatial perception and reasoning & Ego                              & $\triangle$ & $\checkmark$ & $\sim$7M \\
        \hline
        Thinking in Space \cite{ThinkingSpace2024} & Video reasoning and memory & Allo, Ego        & $\times$ & $\times$ & $\sim$5K \\
        \hline
        Perspective Aware \cite{PerspectiveAware2025} & Abstract perspective change & Allo, Ego    & $\checkmark$ & $\times$ & - \\
        \hline
        SPHERE \cite{SPHERE2024} & Spatial perception and hierarchical reasoning & Allo, Ego       & $\triangle$ & $\times$ & $\sim$2K \\
        \hline
        ViewSpatial-Bench \cite{Li2025ViewSpatialBench} & Cross-view spatial reasoning & Allo, Ego & $\checkmark$ & $\times$ & $\sim$5K \\
        \hline
        VLMD4 \cite{VLM4D2025} & Video spatial reasoning & Allo, Ego                               & $\times$ & $\checkmark$ & $\sim$1K \\
        \hline
        AlloEgo-View (ours) & Reference-frame reasoning & Allo, Ego                                & $\checkmark$ & $\checkmark$ & $\sim$4K \\
        \hline
    \end{tabular}
}    
    \vspace{3mm}
    \captionsetup{labelfont={color=mycaptionblue,bf}, textfont={color=mycaptionblue}}
    \caption{Comparison of spatial understanding datasets. 
    MCAF denotes ``multi-choice answer-free queries'' (e.g., \cite{ThinkingSpace2024} requires verbose queries with predefined options). 
    RFF denotes ``reference-frame-free queries'' (e.g., \cite{Li2025ViewSpatialBench} requires explicit reference-frame specification). 
    AlloEgo-View supports natural queries such as ``Where is the refrigerator?'' ($\checkmark$ = yes, $\times$ = no, $\triangle$ = partial).}

\label{tab:spatial_reasoning_comparison}
\vspace{-5mm}
\end{table}

\subsection{Allocentric and Egocentric Perspectives}

\textbf{Egocentric} perspective describes spatial relations from the observer's viewpoint, while \textbf{allocentric} perspective describes them relative to surrounding objects, independent of the observer.
Cognitive studies suggest that humans naturally switch between these perspectives depending on context, task demands, and communicative efficiency~\cite{Orti2025, Alexander2023}.  
However, current VLMs often struggle to robustly handle such variation.  
Recent datasets, such as \textit{ViewSpatial-Bench}~\cite{Li2025ViewSpatialBench} and Spatial-Comfort~\cite{Liu2024SpatialFoRAmbiguities}, highlight the importance of perspective shifts but remain limited in design.  
For instance, they rely on explicit perspective markers in the input (e.g., “from the viewer’s perspective ...”) or adopt multiple-choice formats---both of which deviate from natural language usage, where speakers typically convey spatial relations without verbose reference-frame indicators.

As a result, VLMs often generate inconsistent or even contradictory outputs when faced with short, naturally ambiguous queries.  
This limitation underscores the need for models that can automatically disambiguate perspective-dependent descriptions and provide stable interpretations across both egocentric and allocentric frames---without relying on explicit cues from users. 
A comparison of representative datasets is presented in Table~\ref{tab:spatial_reasoning_comparison}, showing that our AlloEgo-View enables more natural interactions.


\section{AlloEgo-View: A New View-Specific Dataset}
\label{section:AlloEgo-View datyaset}

\begin{wrapfigure}{r}{0.55\textwidth}
    \centering
    \vspace{-10pt}
    \includegraphics[width=0.54\textwidth]{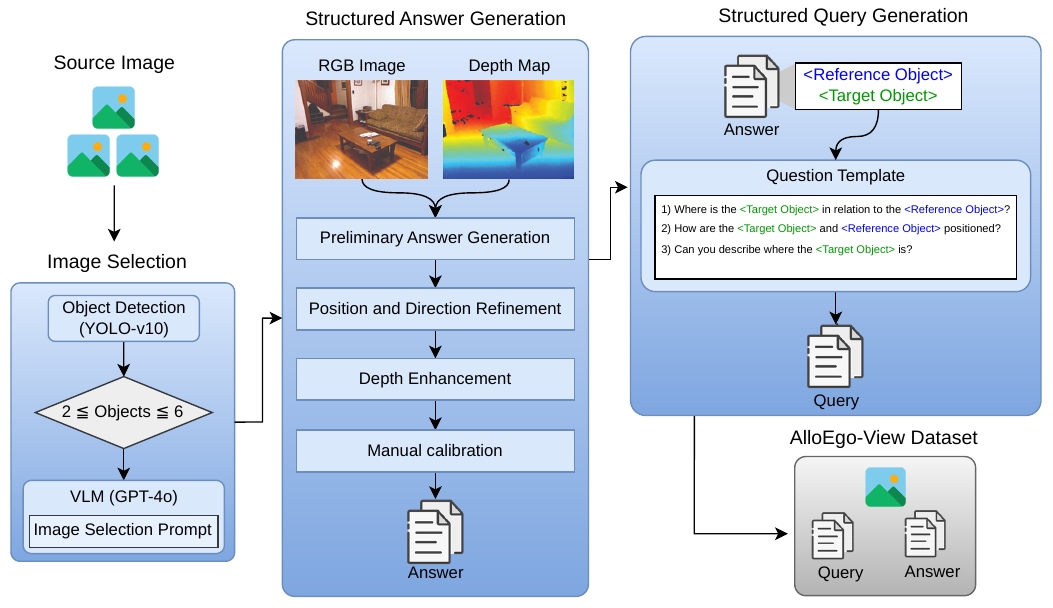}
    \vspace{12pt} 
    \captionsetup{labelfont={color=mycaptionblue,bf}, textfont={color=mycaptionblue}, font=small}
    \caption{AlloEgo-View data curation pipeline. Each triplet consists of an image, a natural but potentially ambiguous query, and an answer that provides detailed egocentric and allocentric view descriptions.}
    \label{fig:dataset_architecture}
    \vspace{-10pt}
\end{wrapfigure}

Spatial reference frame refers to the coordinate system used to specify the position or orientation of an object. 
The AlloEgo-View dataset contains 0.5K manually calibrated samples, 3.5K automatically generated training samples, and 1K manually annotated testing samples. 
Each sample is a view-specific triplet $(\text{image}, \text{query}, \text{answer})$ designed to reduce ambiguity between allocentric and egocentric viewpoints. 
Fig.~\ref{fig:dataset_architecture} illustrates the data curation pipeline.

\subsection{RGB Image Selection}
\label{sec:rgb-select}

We collect images from multiple sources, including \textbf{GQA}~\cite{Hudson2019GQA}, \textbf{SPAR}~\cite{SPAR2025}, \textbf{COCO}~\cite{lin2014coco}, and \textbf{NYU Depth V2}~\cite{silberman2012nyuv2}.  
These images undergo a two-step screening process to ensure suitable reference-frame reasoning.

\textbf{Step 1.} 
Each RGB image is processed using \textbf{YOLOv10}~\cite{wang2024yolov10} for object detection, unless object annotations are already provided by the dataset.  
An image is retained only if it contains between two and six distinct objects.

\textbf{Step 2.} 
Each image is further evaluated using \textit{GPT-4o}.  
A screening prompt determines whether the image contains two to six distinct objects and whether the scene is relatively clean and uncluttered.  
The response is binary, and only images that pass the screening are retained.  
Prompt details are provided in the Supplementary Material.


\subsection{Structured Spatial Answer Generation}
\label{sec:answer-gen}

\begin{wrapfigure}{r}{0.5\textwidth}
    \centering
    \captionsetup{labelfont={color=mycaptionblue,bf}, textfont={color=mycaptionblue}}
   
    \begin{minipage}{\linewidth}
        \centering
        \vspace{-12pt}
        \includegraphics[width=\linewidth]{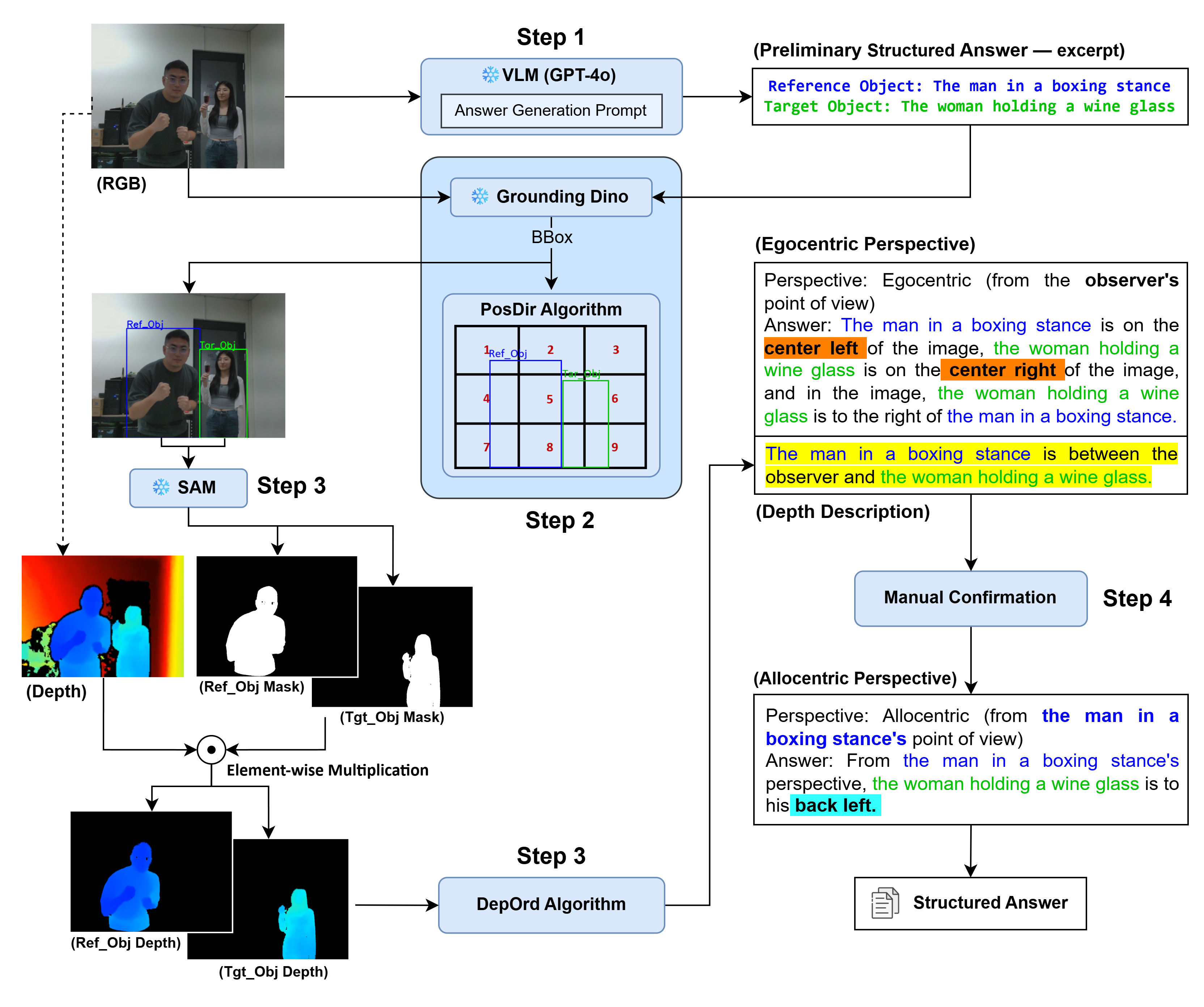}
        \vspace{-14pt}
        \caption*{\small \color{mycaptionblue} (a)}
        \makeatletter
        \edef\@currentlabel{3a}
        \makeatother
        \label{fig:3a}
    \end{minipage}

    \vspace{1pt} 

    \begin{minipage}{\linewidth}
        \centering
        \includegraphics[width=\linewidth]{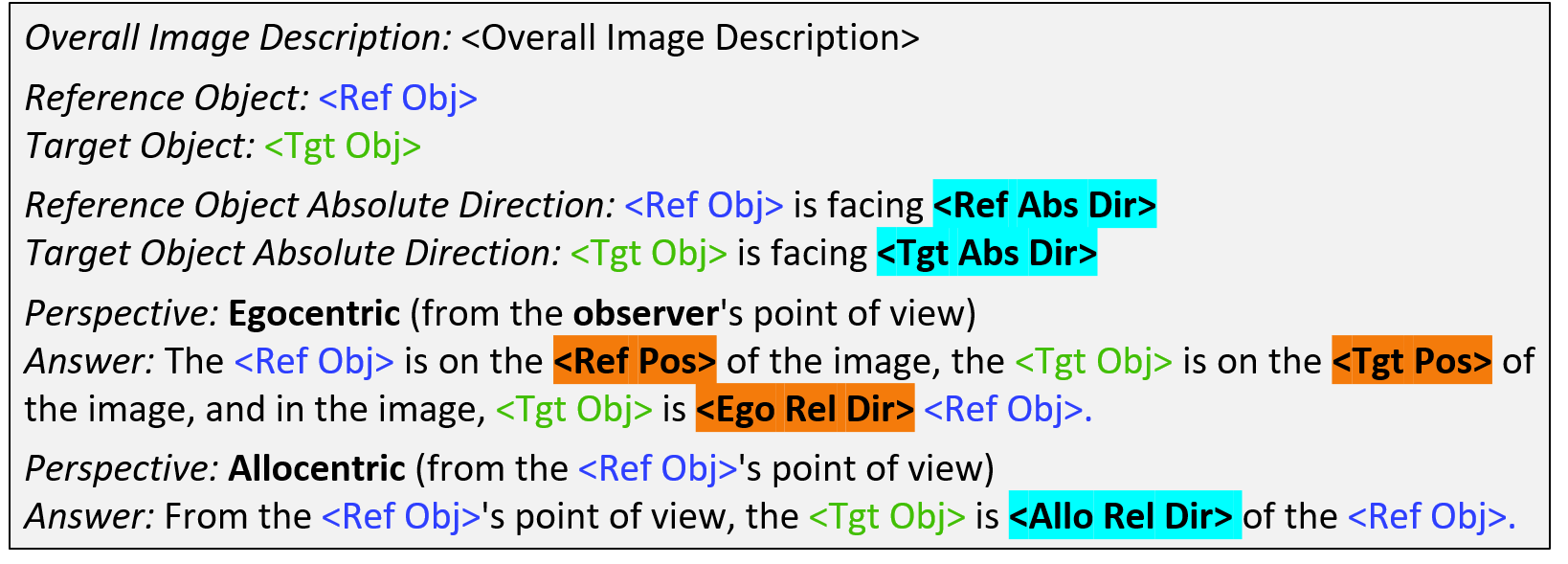}
        \vspace{-6pt}
        \caption*{\small \color{mycaptionblue} (b)}
        \makeatletter
        \edef\@currentlabel{3b}
        \makeatother
        \label{fig:3b}
    \end{minipage}

    \vspace{8pt}
    \caption{Structured spatial answer generation: (a) overall workflow and (b) standardized answer format.}
    \label{fig:3}
    \vspace{-5pt}
\end{wrapfigure}

Next, we describe the workflow for generating structured spatial answers from the retained images (Fig.~\ref{fig:3a}). 
We first construct an initial answer framework (Fig.~\ref{fig:3b}), which is then refined by incorporating allocentric and egocentric view information.



\textbf{Step 1: Preliminary Answer Generation.}  
Each image is processed by GPT-4o (prompt provided in the Supplementary Material) to generate an initial structured answer containing:
\begin{enumerate}[noitemsep]
\item Overall scene description
\item Reference object \texttt{<Ref Obj>} and target object \texttt{<Tgt Obj>}
\item Absolute directions of the reference object \texttt{<Ref Abs Dir>} and target object \texttt{<Tgt Abs Dir>}
\item Egocentric view description
\item Allocentric view description
\end{enumerate}

The output structure is illustrated in Fig.~\ref{fig:3b}.  
The two objects, \texttt{<Ref Obj>} and \texttt{<Tgt Obj>}, are automatically selected by GPT-4o according to the prompt.  
Their absolute directions, \texttt{<Ref Abs Dir>} and \texttt{<Tgt Abs Dir>}, are also determined by GPT-4o.  
In the egocentric view description, the first two statements specify the objects' positions, \texttt{<Ref Pos>} and \texttt{<Tgt Pos>}, within the image, while the third statement describes the relative relation, \texttt{<Ego Rel Dir>}, between the two objects from the observer’s viewpoint.  
In the allocentric view description, the previous two position statements need not be repeated; instead, a single statement specifies the reference frame, \texttt{<Ref Obj>}, and their relative direction, \texttt{<Allo Rel Dir>}.
The rationale behind designing this structured format will be explained in the subsequent experiments.

According to our experiments, most VLMs tend to produce multiple errors in the above descriptions.  
To correct these errors, \texttt{<Ref Pos>}, \texttt{<Tgt Pos>}, and \texttt{<Ego Rel Dir>} in the egocentric view description are refined in Step~2.  
Step~3 further enhances the description by incorporating depth information, thereby capturing the spatial front--back relations between objects.  
Remaining errors are manually corrected in Step~4, primarily involving updates to \texttt{<Ref Abs Dir>}, \texttt{<Tgt Abs Dir>}, and the allocentric view description.

In the example shown in Fig.~\ref{fig:3}, the orange-highlighted parts are corrected in Step~2, the yellow-highlighted parts are refined in Step~3, and the blue-highlighted parts are manually adjusted in Step~4. 
The complete final answer is provided in the Supplementary Material.



\textbf{Step 2: Position and Direction Refinement.}  
To calibrate \texttt{<Ref Pos>}, \texttt{<Tgt Pos>}, and \texttt{<Ego Rel Dir>}, we first apply \textbf{Grounding DINO}~\cite{liu2023groundingdino} using \texttt{<Ref Obj>}, \texttt{<Tgt Obj>}, and the image as inputs.  
Grounding DINO identifies the instance-level grounding of each object, and its outputs are the corresponding bounding boxes.

Next, we propose the \textbf{PosDir} algorithm, which determines the absolute and relative positions of the two objects in the image from an egocentric viewpoint.
\begin{enumerate}[noitemsep]
\item 
Evenly partition the image into a $3\times3$ grid.  
Textual descriptions \texttt{<Ref Pos>} and \texttt{<Tgt Pos>} are generated for \texttt{<Ref Obj>} and \texttt{<Tgt Obj>} based on the overlap between their bounding boxes and the grid regions.
\item 
The textual description \texttt{<Ego Rel Dir>} is generated according to the relative positions of the centers of the two bounding boxes.
\end{enumerate}
These descriptions are then used to update the initial answer generated in Step~1.  
In Fig.~\ref{fig:3a}, the orange-highlighted segments indicate the updated egocentric view description.  
The complete algorithm is provided in the Supplementary Material.


\textbf{Step 3: Depth Enhancement.}
As an egocentric view involves two objects and the observer, the relative depth ordering is critical.
To incorporate depth information, we first extract the segmentation maps of \texttt{<Ref Obj>} and \texttt{<Tgt Obj>} using \textbf{Segment Anything Model (SAM)}~\cite{kirillov2023sam}, where the input bounding boxes are provided by Grounding DINO from Step 2.
Second, we compute the object's depth maps by performing element-wise multiplication between each object's segmentation map and the image depth map:
$
SAM(\texttt{<Ref Obj>}/\texttt{<Tgt Obj>}) \odot Depth(\text{Image})
$.
Third, we calculate the mean depth of each object based on its depth centroid. 
We then propose the \textbf{DepOrd} algorithm to determine their relative depth ordering. 
By comparing the mean depths, one of the following three relations is obtained: 
``\texttt{<Ref Obj>} is deeper,'' ``\texttt{<Tgt Obj>} is deeper,'' or ``no depth ordering.''

For example, in Fig.~\ref{fig:3a}, the man is closer to the observer, forming a front–back relationship with the woman. Therefore, an additional description (yellow-highlighted parts), “The man in a boxing stance is between the observer and the woman holding a wine glass,” is included. The complete algorithm is provided in the Supplementary Material.


\textbf{Step 4: Manual calibration.}  
We address errors through manual review. This primarily involves updates to \texttt{<Ref Abs Dir>}, \texttt{<Tgt Abs Dir>}, and \texttt{<Allo Rel Dir>}. Totally 0.5K answers are produced. 
We note that \texttt{<Allo Rel Dir>} is particularly prone to errors, as it heavily depends on the facing direction of \texttt{<Ref Obj>}, which necessitates additional manual effort.
In Fig.~\ref{fig:3a}, the blue-highlighted parts are updated.

\subsection{Structured Spatial Query Generation}
\label{sec:question-gen}

To enable AlloEgo-VLM to respond naturally during inference, we associate each above answer with a structured spatial question.
We define a set of 17 question templates to capture a wide variety of user queries, ranging from direct relational questions to more conversational descriptions. 
This ensures linguistic diversity while maintaining semantic consistency.
Given any pair (\texttt{<Ref Obj>}, \texttt{<Tgt Obj>}), we show two representative question examples:
\begin{enumerate}[label=\alph*), noitemsep]
\item 
What is the relationship between the \texttt{<Tgt Obj>} and the \texttt{<Ref Obj>}?
\item 
Could you describe where the \texttt{<Tgt Obj>} is located?
\end{enumerate}

During dataset construction, one random template is selected for each question. The complete list of 17 templates is provided in the Supplementary Material.


\begin{wrapfigure}{r}{0.5\textwidth}
    \centering
    \vspace{-22pt} 
    \captionsetup{labelfont={color=mycaptionblue,bf}, textfont={color=mycaptionblue}, }
    \includegraphics[width=0.49\textwidth]{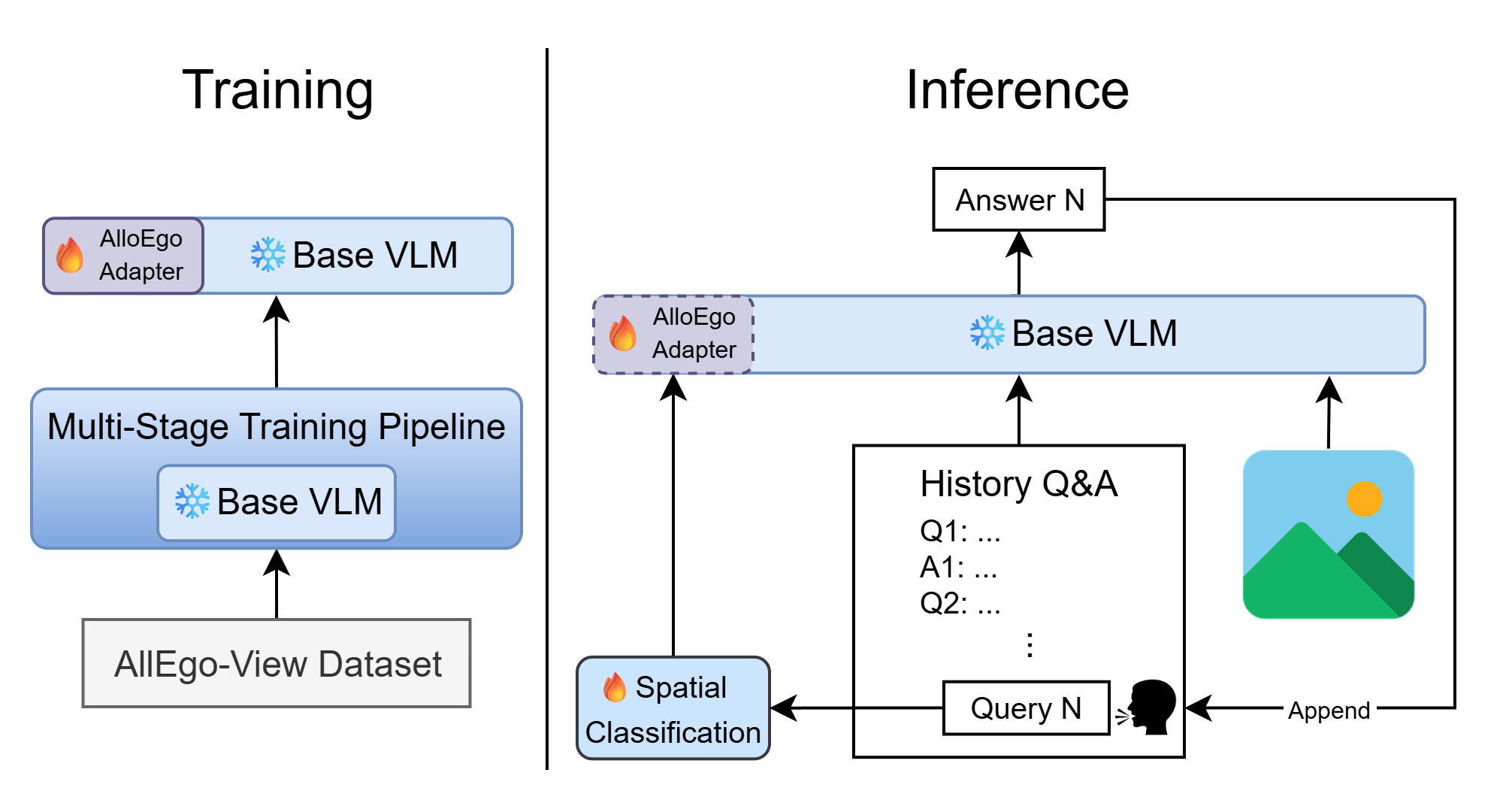}
    \vspace{7pt} 
    \caption{AlloEgo-VLM Framework.}
    \label{fig:method_overall}
    \vspace{-7pt} 
\end{wrapfigure}


\section{AlloEgo-VLM for Reference--Frame Disambiguation}
\label{sec:AlloEgo-VLM}

The proposed \textbf{AlloEgo-VLM} is designed to answer not only questions with reference-frame ambiguity, but also general non-spatial reasoning questions, while supporting multi-round conversations.  
The overall framework is illustrated in Fig.~\ref{fig:method_overall}.

We adopt a base VLM and fine-tune it with an \textbf{AlloEgo-Adapter} that follows our structured spatial reasoning format. 
Although the initial AlloEgo-View dataset is small in scale, it enables the use of quantization, LoRA-based~\cite{Dettmers2023QLoRA} supervised fine-tuning (SFT), and bootstrapped data generalization to progressively expand the dataset and train a robust model.  

To address view-specific queries, we train a lightweight \textbf{Spatial Query Classifier} to determine the query type.  
If a query is identified as view-specific, the model activates the AlloEgo-Adapter for structured spatial processing; otherwise, the base VLM directly generates the response.  
This design enables efficient handling of both spatial and non-spatial tasks.

We adopt \textit{Qwen~2.5-VL-7B}~\cite{qwen2025qwen2_5_vl_7b}, \textit{Llama~3.2-V-11B}~\cite{meta2024llama3_2_v}, and \textit{Gemma~3-V-4B}~\cite{deepmind2025gemma3_v} as the backbone VLMs of AlloEgo-VLM.  
The lightweight classifier is trained on a single NVIDIA RTX~3090 GPU.

\subsection{Multi-stage Training of AlloEgo-Adapter}
\label{sec:4a}

\begin{wrapfigure}{r}{0.5\textwidth}
    \centering
    \vspace{-10pt}
    \captionsetup{labelfont={color=mycaptionblue,bf}, textfont={color=mycaptionblue}}
    \includegraphics[width=0.49\textwidth]{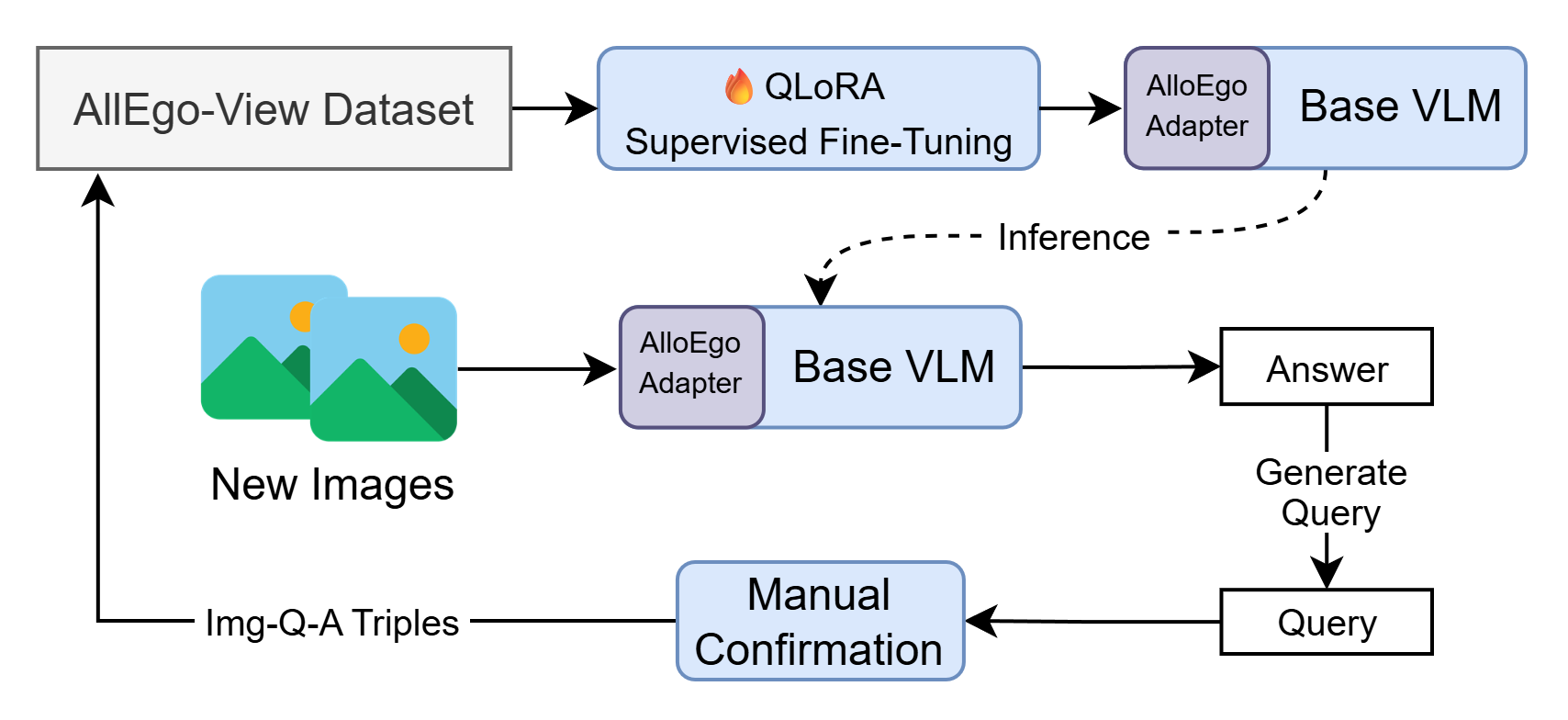}
    \vspace{8pt} 
\caption{Multi-stage training pipeline of AlloEgo-Adapter and progressive expansion of AlloEgo-View.}
    \label{fig:DPO_two_step_training}
    \vspace{-5pt} 
\end{wrapfigure}

From a frozen base VLM, we train an AlloEgo-Adapter in an iterative manner. During the multi-stage training, we also expand the initial AlloEgo-View dataset from a scale of 0.5K to 4K, inspired by \textbf{DPO} \cite{Rafailov2023DPO}.
The pipeline is shown in Fig.~\ref{fig:DPO_two_step_training}.


\textbf{Step 1: AlloEgo-Adapter Fine-Tuning.} 
We perform QLoRA-based~\cite{Dettmers2023QLoRA} SFT on the base VLM using the AlloEgo-View dataset, resulting in \textbf{AlloEgo-Adapter}, which aligns the base VLM with our structured spatial reasoning format.

\textbf{Step 2: Bootstrapped Data Generation.} 
Following Sec.~\ref{sec:rgb-select}, we collect additional images and feed them into the initial model equipped with the AlloEgo-Adapter trained in Step~1. 
Because Step~1 employs full-format SFT, the output structure is strongly constrained, enabling the model to generate structured answers directly from images without explicit input queries. 
We then generate the corresponding questions using the procedure in Sec.~\ref{sec:question-gen}, producing additional $(\text{image}, \text{query}, \text{answer})$ triplets consistent with the AlloEgo-View format.

\textbf{Step 3: Manual Confirmation and Iterative Refinement.} 
The generated triplets undergo manual verification to ensure correctness and natural language quality. 
The validated triplets are then added to the AlloEgo-View dataset. 
Finally, Steps~1--3 are iteratively repeated until convergence.

In summary, the pipeline progressively expands AlloEgo-View through bootstrapped generation and iteratively fine-tunes AlloEgo-Adapter for robust spatial reasoning. 
In practice, five iterations are sufficient for stable performance.


\subsection{View-Specific Spatial Query Classifier}
\label{sec:spatial_cls}

\begin{wrapfigure}{r}{0.5\textwidth}
    \centering
    \vspace{-13pt}
    \captionsetup{labelfont={color=mycaptionblue,bf}, textfont={color=mycaptionblue}}
    \includegraphics[width=0.49\textwidth]{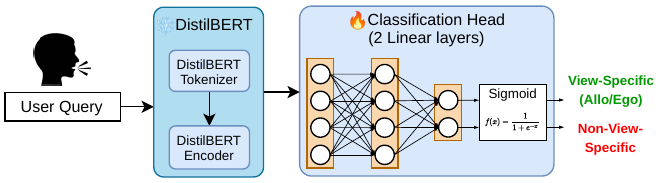}
    \vspace{8pt}
\caption{Spatial Query Classifier for filtering and categorizing spatial-related user queries.}
    \label{fig:Spatial_Classification}
    \vspace{-10pt} 
\end{wrapfigure}

Due to the rigid SFT output format, the adapter-enhanced VLM tends to favor spatial responses. 
To distinguish view-specific (egocentric/allocentric) queries from general ones, we train a lightweight classifier, as illustrated in Fig.~\ref{fig:Spatial_Classification}. 
The query is tokenized using a frozen \textbf{DistilBERT tokenizer}~\cite{sanh2019distilbert} and encoded by a frozen \textbf{DistilBERT encoder} to obtain a latent representation while preserving generalizability. 
A lightweight classification head with two linear layers then predicts whether the query involves spatial-direction reasoning. 
We adopt DistilBERT due to its efficiency, compact size, and strong semantic encoding capability.

To train the classifier, we construct a question-only dataset, \textbf{AlloEgo-QO}, containing approximately 10K queries without answers.  
It includes 4K spatial queries from AlloEgo-View and 6K non-spatial queries collected from online sources (e.g., mathematics, finance, and daily tasks).  
The former are labeled as positive (view-specific) and the latter as negative (non-view-specific).  
The train--test split ratio is 8:2.
During inference, each question is first processed by the Spatial Classifier. If it corresponds to a spatial-direction reasoning task, both the base VLM and AlloEgo-Adapter are invoked; otherwise, only the base VLM is used.


\subsection{Support for Multi-round Dialogue}
\label{sec:4c}

The framework in Fig.~\ref{fig:method_overall} supports multi-round dialogue. 
The conversation history, query image, and current question are fed into the VLM in an autoregressive manner. 
The invocation of AlloEgo-Adapter is determined by the Spatial Query Classifier. 
The generated answer, together with the current question, is appended to the dialogue history to preserve contextual information. 
Subsequent interactions are processed iteratively in the same manner.

Since our ultimate goal is robotic deployment, a new dialogue session is initiated whenever the robot observes a new scene. 
Therefore, in our multi-round setting, the dialogue history contains only previous questions and answers, while the query image is not re-input in subsequent rounds.


\section{Experiments}

\subsection{View-Specific Comparisons}
\label{sec:view-specific}

\begin{wraptable}{r}{0.48\linewidth}
\centering
\footnotesize
\renewcommand{\arraystretch}{1.05}

\begin{tabular}{lccc}
\toprule
\textbf{Method} & \textbf{Params} & \textbf{VAR $\downarrow$} & \textbf{VSA $\uparrow$} \\
\midrule
Qwen 2.5-VL   & 7B  & 94.568\% & 50\% \\
Llama 3.2-V   & 11B & 100\%    & -- \\
Gemma 3-V     & 4B  & 99.753\% & 100\% \\
GPT-4o        & --  & 99.259\% & 66.666\% \\
GPT-4o mini   & --  & 99.012\% & 75\% \\
\bottomrule
\end{tabular}

\vspace{3mm}
\captionsetup{labelfont={color=mycaptionblue,bf}, textfont={color=mycaptionblue}}
\caption{Perspective tests of strong VLMs.}
\label{tab:Ambiguity_rate}
\vspace{-3mm}
\end{wraptable}

We integrate different base VLMs into our framework and compare against several state-of-the-art models, including \textbf{GPT-4o}, \textbf{GPT-4o mini}, \textbf{Llama 3.2-V}, \textbf{Gemma 3-V}, and variants of \textbf{Qwen 2.5-VL}. Our evaluation focuses on view awareness and reference-frame disambiguation capability.

\textbf{Allocentric-Egocentric View Awareness Test.}
\label{sec:Allocentric-Egocentric View Awareness Test}
We select the questions from AlloEgo-View that are spatially ambiguous, meaning no reference frame or perspective is provided. 
By feeding only these questions to a model without any view-specific format constraints, we aim to assess whether the model possesses sufficient view awareness.

Table~\ref{tab:Ambiguity_rate} reports two metrics. 
The \textbf{View Ambiguity Rate (VAR)} evaluates whether a model's response introduces confusion (prompt available in the Supplementary Material).
VAR measures the proportion of cases where a query lacks reference frame information and the model also omits it, leading to ambiguous interpretations. 
All models exhibit high ambiguity: \textit{Qwen2.5-VL (7B)} 94.57\%, \textit{Llama3.2-V (11B)} 100\%, and \textit{GPT-4o} variants above 99\%. 
These results indicate that current models cannot reliably resolve spatial relations without explicit reference frames.

The \textbf{View-Specific Accuracy (VSA)} measures how often a model gives the correct answer when its response explicitly includes a reference frame. Because the evaluation set is small, each case is manually verified. Most models show inconsistent performance in this aspect, highlighting the importance of AlloEgo-View, which provides view-specific QA.

\textbf{Allo-Egocentric View Disambiguation Test.}
Table~\ref{tab:model_comparison} presents a comprehensive comparison of \textit{Llama~3.2-V}, \textit{Gemma~3-V}, variants of \textit{Qwen~2.5-VL}, \textit{GPT-4o}, and \textit{GPT-4o~mini}. 
The first three are further extended with the AlloEgo-Adapter. 
The test dataset consists of 1K manually annotated samples (refer to Sec. \ref{section:AlloEgo-View datyaset}).
Three evaluation settings are reported:
\begin{enumerate}[noitemsep]
\item 
\textbf{View-Specific Format-Only Prompt (VS-Format):} 
The model receives only the query and the expected output format, without any additional guidance on spatial semantics or structured reasoning.
\item 
\textbf{View-Specific Textbook-Level Prompt (VS-Textbook):} 
The model receives a detailed prompt that instructs it to distinguish between allocentric and egocentric descriptions, avoid ambiguous or contradictory answers, and follow structured spatial reasoning.
\item 
\textbf{Proposed SFT:} 
A backbone model adapted through SFT (Sec.~\ref{sec:AlloEgo-VLM}).
\end{enumerate}

We evaluate the key entries highlighted in Fig.~\ref{fig:3b} and report: 
\textbf{AD\_RO} (Reference Object Absolute Direction), \textbf{AD\_TO} (Target Object Absolute Direction), \textbf{Ego} (egocentric), and \textbf{Allo} (allocentric). 
Each entry in the test dataset is annotated with a ground truth. 
We use \textit{GPT-4o} to score answer–ground-truth similarity from 1 to 10 (see the Supplementary Material).

\begin{table}[t]
    \centering
    \footnotesize 
    \renewcommand{\arraystretch}{1.3}
    \setlength{\tabcolsep}{3pt} 
        \begin{tabular}{lccccc}
        \textbf{Method} & \textbf{Params} & \textbf{AD\_RO↑} & \textbf{AD\_TO↑} & \textbf{Ego↑} & \textbf{Allo↑} \\
        \hline
        VS-Format (Qwen 2.5-VL)   & 7B  & 4.08 & 3.62 & 4.51 & 2.65 \\
        VS-Textbook (Qwen 2.5-VL) & 7B  & 5.22 & 4.61 & 5.39 & 4.16 \\
        \rowcolor{gray!15}
        AlloEgo-VLM (Qwen 2.5-VL)      & 7B  & \textbf{7.94} & \textbf{8.04} & \textbf{8.19} & \textbf{6.25} \\
        \hline
        VS-Format (Llama 3.2-V)   & 11B & 3.82 & 3.53 & 4.36 & 3.74 \\
        VS-Textbook (Llama 3.2-V) & 11B & 4.95 & 4.56 & 5.03 & 3.38 \\
        \rowcolor{gray!15}
        AlloEgo-VLM (Llama 3.2-V)      & 11B & \textbf{7.92} & \textbf{8.17} & \textbf{8.28} & \textbf{6.72} \\
        \hline
        VS-Format (Gemma 3-V)     & 4B  & 3.59 & 3.20 & 1.84 & 3.94 \\
        VS-Textbook (Gemma 3-V)   & 4B  & 4.75 & 4.50 & \textbf{4.89} & 3.86 \\
        AlloEgo-VLM (Gemma 3-V)        & 4B  & \textbf{5.38} & \textbf{5.76} & 4.36 & \textbf{4.30} \\
        \hline
        VS-Format (GPT-4o mini)   & -   & 4.40 & 4.16 & 5.67 & 4.43 \\
        VS-Textbook (GPT-4o mini) & -   & \textbf{6.32} & \textbf{6.41} & \textbf{7.02} & \textbf{5.27} \\
        \hline
        VS-Format (GPT-4o)        & -   & 4.54 & 4.56 & 5.95 & 5.05 \\
        \rowcolor{gray!15}
        VS-Textbook (GPT-4o)      & -   & \textbf{7.25} & \textbf{7.42} & \textbf{7.34} & \textbf{5.91} \\
        \end{tabular}%
    
    \vspace{8pt}
    \captionsetup{labelfont={color=mycaptionblue,bf}, textfont={color=mycaptionblue}}
    \caption{Comparison of disambiguation capability.}
    \label{tab:model_comparison}
    \vspace{-10pt}
\end{table}

From Table~\ref{tab:model_comparison}, several key observations can be drawn:
\begin{itemize}[noitemsep]
\item 
Each model shows clear progressive improvement: VS-Format yields moderate performance, VS-Textbook brings substantial gains across all metrics, and the full SFT setting achieves the best results (bolded). This trend underscores the contribution of AlloEgo-VLM. \textit{Gemma 3-V}, due to its smaller parameter size, scores slightly lower on the Ego metric in AlloEgo-VLM compared to VS-Textbook.
\item 
When comparing the gray-shaded rows, it is notable that \textit{Qwen 2.5-VL (7B)} and \textit{Llama 3.2-V (11B)}, both using our AlloEgo-VLM method, outperform the strongest proprietary baseline, VS-Textbook (\textit{GPT-4o}).

\item 
The superior performance of our SFT approach stems from explicit spatial encoding in AlloEgo-View, iterative fine-tuning with human verification, and a structured output format that improves generalization across base models.
\end{itemize}

\subsection{Performance of the Spatial Query Classifier}

The spatial classifier plays a key role in preventing AlloEgo-VLM from overfitting to spatial questions.  
We evaluate it on the AlloEgo-QO test set, with results summarized in Table~\ref{tab:validation_Metrics_for_Spatial_Classification}.  
We conduct the same experiment on the state-of-the-art VLMs via prompting.
Our model achieves perfect accuracy (100\%) while maintaining a lightweight design (6M parameters, 0.25GB) and extremely fast inference speed ($\approx$2ms).  
Since this is a relatively simple task, \textit{Qwen 2.5-VL} and \textit{Llama 3.2-V} also achieve accuracies above 99\%, while \textit{Gemma 3-V} lags behind on this binary decision task.  
Thus, our lightweight spatial classifier effectively serves its purpose in our setting.

\begin{wraptable}{l}{0.45\linewidth}
\centering
\footnotesize
\renewcommand{\arraystretch}{1.15}
\setlength{\tabcolsep}{2.5pt}

\begin{tabular}{lcccc}
\toprule
\textbf{Method} & \textbf{Para.} & \textbf{Size} & \textbf{Inf.} & \textbf{Acc.} \\
\midrule
Spatial Classifier & 6M  & 0.25GB & $\approx$2ms   & 100\% \\
Qwen 2.5-VL        & 7B  & 7GB    & $\approx$260ms & 99.53\% \\
Llama 3.2-V        & 11B & 11GB   & $\approx$700ms & 99.71\% \\
Gemma 3-V          & 4B  & 4GB    & $\approx$240ms & 91.24\% \\
\bottomrule
\end{tabular}
\vspace{5mm}
\captionsetup{labelfont={color=mycaptionblue,bf}, textfont={color=mycaptionblue}}
\caption{Capability for recognizing spatial queries.}
\label{tab:validation_Metrics_for_Spatial_Classification}
\vspace{3mm}
\end{wraptable}

\begin{wrapfigure}{r}{0.5\textwidth}
    \centering
    \vspace{-130pt}
    \captionsetup{labelfont={color=mycaptionblue,bf}, textfont={color=mycaptionblue}}

    \begin{minipage}{0.23\textwidth}
        \centering
        \includegraphics[width=\linewidth]{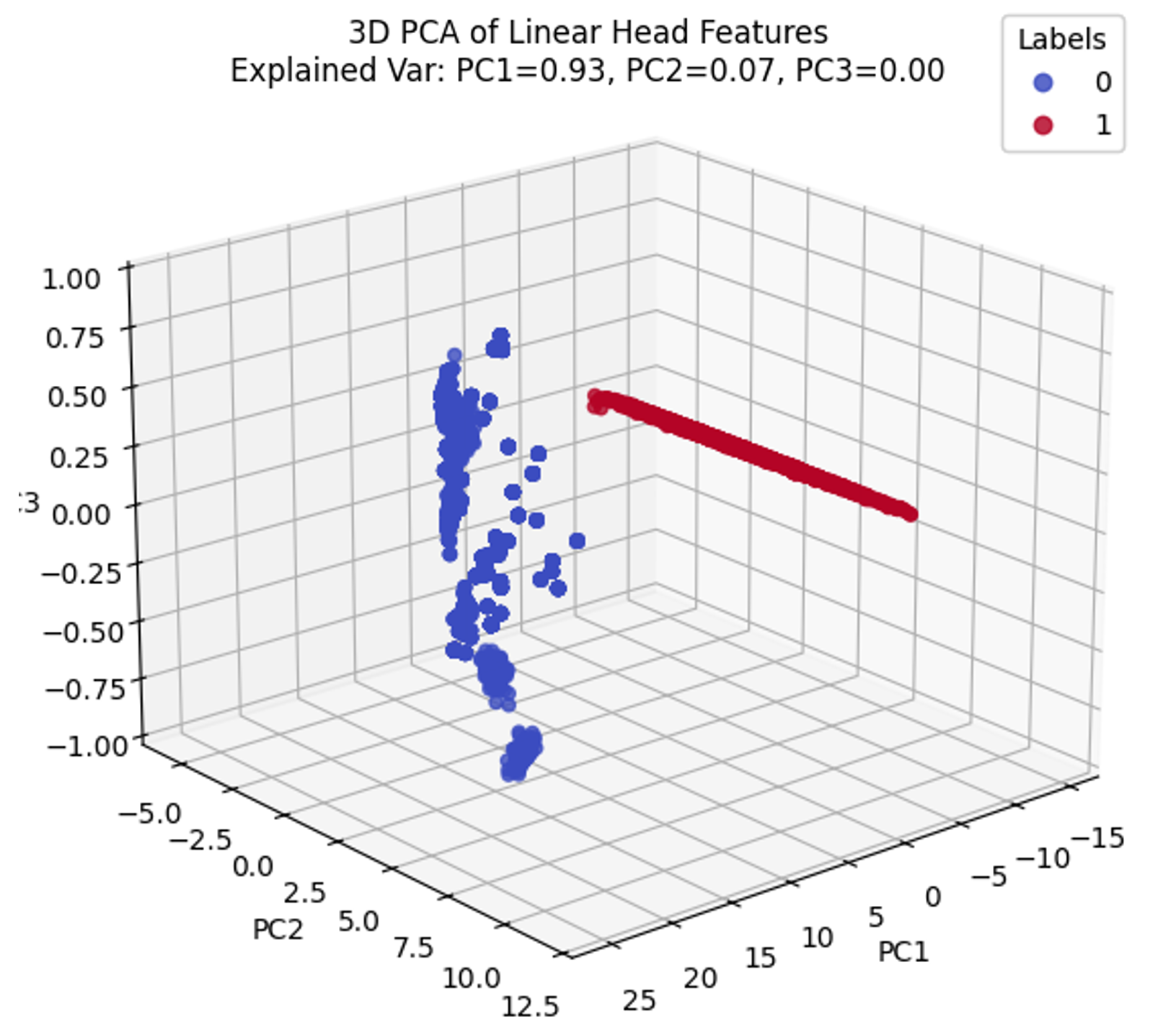}
        \vspace{-10pt}
        \caption*{\scriptsize \color{mycaptionblue} (a) Training set}
        \makeatletter
        \edef\@currentlabel{\the\numexpr\value{figure}+1\relax(a)}
        \makeatother
        \label{fig:train_pca}
    \end{minipage}
    \hfill 
    \begin{minipage}{0.23\textwidth}
        \centering
        \includegraphics[width=\linewidth]{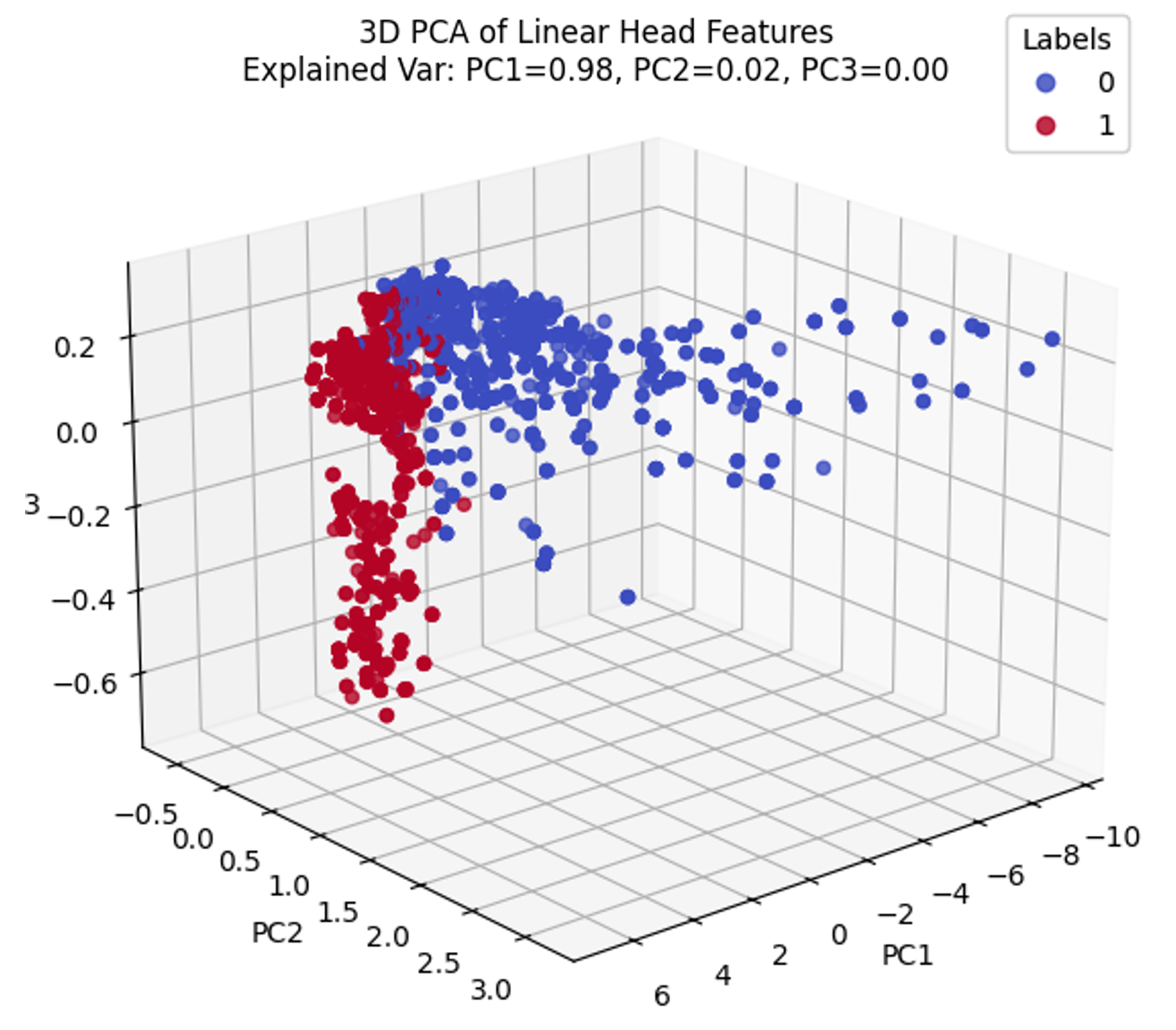}
        \vspace{-5pt}
        \caption*{\scriptsize \color{mycaptionblue} (b) Test set}
        \makeatletter
        \edef\@currentlabel{\the\numexpr\value{figure}+1\relax(b)}
        \makeatother
        \label{fig:test_pca}
    \end{minipage}

    \vspace{7pt}
\caption{3D PCA projection of features encoded by the Spatial Query Classifier.}
    \label{fig:train_test_pca}
    \vspace{-15pt}
\end{wrapfigure}

To examine the learned representations, we visualize the final M3 layer features using \textbf{3D PCA}~\cite{jolliffe2002pca}.  
Fig.~\ref{fig:train_test_pca} shows clear separation between the two classes in both training and unseen test data, confirming the learned discriminative features.


\subsection{Training Performance of AlloEgo-Adapter}

\begin{wrapfigure}{l}{0.5\textwidth}
    \centering
    \captionsetup{labelfont={color=mycaptionblue,bf}, textfont={color=mycaptionblue}}
    \includegraphics[width=0.48\textwidth]{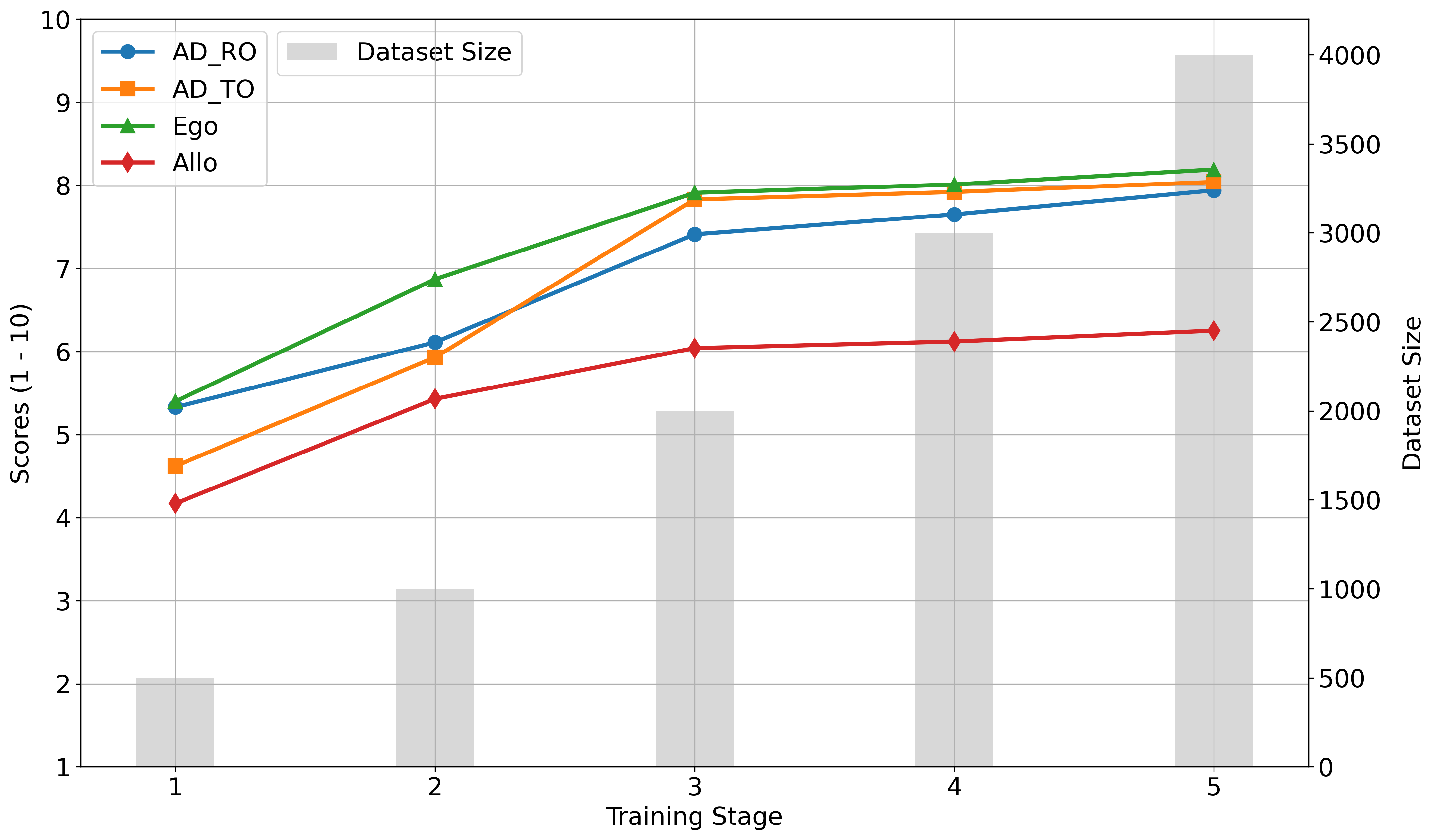}
    \vspace{8pt}
\caption{Progressive training of AlloEgo-Adapter on \textit{Qwen~2.5-VL-7B}. Line: performance; bars: dataset size.}
\label{fig:multi_stage}
    \vspace{-20pt}
\end{wrapfigure}

AlloEgo-Adapter was trained progressively in a multi-stage manner, with each stage generating additional view-specific data for fine-tuning.  
We selected \textit{Qwen~2.5-VL-7B} as the backbone, which offers a favorable balance between capacity and performance.

Fig.~\ref{fig:multi_stage} shows its performance trajectory.
The model improves steadily in the early stages and begins to plateau after Stage~3.  
The bar chart illustrates the dataset growth during bootstrapped data generation. 
At Stage~5, where the training dataset reaches 4K samples, performance stabilizes.  
These results indicate that AlloEgo-VLM can be effectively trained with a relatively small dataset of 4K samples, achieving performance levels that approach or surpass 8 (out of 10) across the majority of evaluation metrics.




\subsection{Embodied Deployment for Open-Ended Spatial Grounding}
\label{sec:embodied_deployment}




\begin{table}[h]
\centering
\footnotesize
\setlength{\tabcolsep}{3.8pt} 
\renewcommand{\arraystretch}{1.3}

\begin{tabular}{l | ccc | ccc}
\toprule
& \multicolumn{3}{c|}{\textbf{AQ (implicit allocentric SR)}} & \multicolumn{3}{c}{\textbf{EQ (explicit egocentric SR)}} \\
\textbf{Scenario <reference obj>} & \textbf{Qwen-For} & \textbf{Qwen-Txt} & \textbf{AlloEgo} & \textbf{Qwen-For} & \textbf{Qwen-Txt} & \textbf{AlloEgo} \\
\midrule
Case 1: Table (Non-directional)   & 1.0 & 1.0 & \textbf{1.0} & 1.0 & 0.9 & \textbf{1.0} \\
Case 2: Sofa (Facing Observer)    & 0.0 & 0.1 & \textbf{1.0} & 0.9 & 1.0 & \textbf{1.0} \\
Case 3: Chair (Facing Observer)   & 0.2 & 0.2 & \textbf{1.0} & 0.8 & 0.9 & \textbf{0.9} \\
Case 4: Chair (Facing Left)       & 1.0 & 0.7 & 0.8 & 0.4 & 0.6 & \textbf{0.8} \\
Case 5: Human (Facing Observer)   & 0.2 & 0.1 & \textbf{0.4} & 0.3 & 0.5 & \textbf{0.5} \\
Case 6: Wheelchair (Facing Away)  & 1.0 & 1.0 & 0.5 & 0.8 & 1.0 & 0.7 \\ 
\midrule
\textbf{Average Success Rate (SR) $\uparrow$} & 0.567 & 0.517 & \textbf{0.783} & 0.700 & \textbf{0.816} & \textbf{0.816} \\
\bottomrule
\end{tabular}

\vspace{3mm} 
\captionsetup{labelfont={color=mycaptionblue,bf}, textfont={color=mycaptionblue}}
\caption{Navigation Success Rate (SR) for embodied search in Isaac Sim experiments. 
\textit{AQ (Ambiguous Queries)} evaluate implicit allocentric transformations with unstated reference frames, while \textit{EQ (Explicit Queries)} serve as an egocentric control benchmark for standard camera-space grounding using the phrase \textit{``from your perspective''}.}
\label{tab:embodied_sr}
\end{table}

To evaluate the practical utility of AlloEgo-VLM in embodied AI, we develop a framework in \textbf{NVIDIA Isaac Sim}. 
We design an open-ended object search task requiring both long-horizon navigation and fine-grained spatial reasoning, following the workflow below:

\begin{enumerate}[noitemsep]
\item 
\textbf{Reference-Centric Navigation.} 
Given a complex spatial instruction (e.g., \textit{``find the plant right of the sofa''}), the robot first uses an LLM planner to navigate toward the designated reference object (\textit{sofa}). 
This step isolates global path planning and focuses the evaluation on local spatial understanding.

\item
\textbf{Perception Triggering.} 
Upon arriving near the \textit{sofa}, the LLM planner issues a \texttt{call\_vlm} command, triggering the robot's onboard RGB camera(s) to actively perceive the surrounding scene.

\item
\textbf{Candidate Object Localization.}  
To address semantic ambiguity, Grounding DINO identifies all candidate \textit{plant} instances in the captured images. Red bounding boxes are overlaid as visual prompts, producing multiple prompted images for VLM reasoning.

\item
\textbf{Spatial Reasoning and Grounding.} 
Each image is queried with: \textit{``Can you describe where the \{target\} in the red box is?''}. 
The generated answers are verified by matching their directional and perspective descriptions against the original instruction. 
The candidate with the highest semantic alignment is selected as the target object.
\end{enumerate}

\begin{wrapfigure}{l}{0.5\textwidth}
    \vspace{-5mm}
    \centering
    \captionsetup{labelfont={color=mycaptionblue,bf}, textfont={color=mycaptionblue}}
    \includegraphics[width=0.49\textwidth]{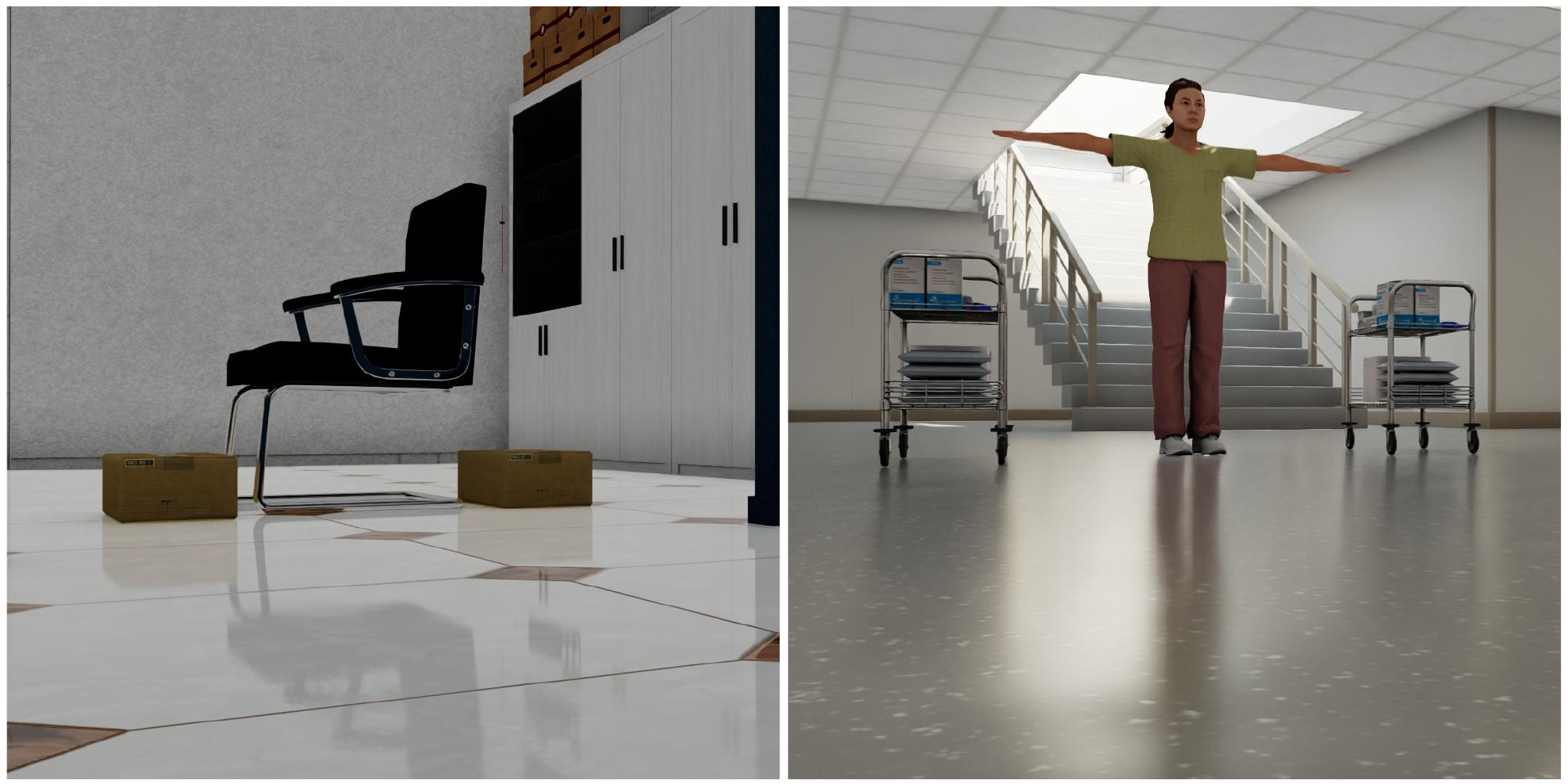}
    \vspace{8pt}
\caption{Example multi-object scenarios in Isaac Sim for embodied object search tasks. 
Boxes are positioned in front of and behind the chair (left), while carts are placed on both sides of the human avatar (right).}
\label{fig:scene}
    \vspace{-10pt}
\end{wrapfigure}

We establish two baselines under different prompting settings for spatial reasoning: 
(i) \textbf{VS-Format (corresponding to Qwen-For in Table~\ref{tab:embodied_sr}):} evaluation without explicit perspective or coordinate-system instructions, using the standard \textit{Qwen~2.5-VL} backbone; 
(ii) \textbf{VS-Textbook (corresponding to Qwen-Txt in Table~\ref{tab:embodied_sr}):} evaluation with zero-shot prompts augmented by textbook-style spatial reasoning demonstrations and structured reference-frame definitions on the \textit{Qwen~2.5-VL} backbone. 
For AlloEgo-VLM, we adopt \textit{Qwen~2.5-VL} as the backbone and apply our reference-frame alignment fine-tuning protocol. 
All methods are integrated into the same robotic pipeline for fair comparison.

    
    

We conduct closed-loop object search experiments in two high-fidelity NVIDIA Isaac Sim environments containing \textit{multiple} similar objects: an office scene and a hospital scene (Fig.~\ref{fig:scene}). 
Within these environments, we design six testing scenarios under two settings: 
(i) \textbf{Non-directional References:} objects without an intrinsic front, where spatial frames depend entirely on the observer (e.g., symmetrical tables); 
(ii) \textbf{Directional References:} objects with clear intrinsic fronts that support allocentric reference frames (e.g., sofas and wheelchairs). 
Table~\ref{tab:embodied_sr} summarizes the six scenarios.
For each scenario, the robot performs a spatial disambiguation task using either ambiguous queries (e.g., \textit{``Find the table right of the wheelchair''}) or explicitly constrained egocentric queries (e.g., \textit{``Find the cart left of the human from your perspective''}). 
Each case is evaluated over $10$ independent trials with randomized starting locations, resulting in diverse stopping positions, viewing distances, and camera angles to rigorously assess reference-frame stability.


We report the \textbf{Navigation Success Rate (SR)} in Table~\ref{tab:embodied_sr}. 
A trial is considered successful only if it satisfies two criteria: 
(i) \textbf{Perceptual Grounding}: the VLM correctly identifies the target object while filtering out distractors through alignment verification; and 
(ii) \textbf{Physical Arrival}: the robot autonomously navigates to within $1.5\,\mathrm{m}$ of the target object. 
Failure to satisfy either criterion is counted as a failure.


The overall SR across all trials yields two key takeaways. 
First, under the ambiguous-query condition, which inherently requires allocentric reference-frame mapping, AlloEgo-VLM achieves a substantial performance gain with an average SR of 0.783, outperforming the baseline Qwen-For (0.567) and the prompted Qwen-Txt (0.517).
Second, the results under the explicit-query control condition demonstrate the architectural stability of our framework. 
When evaluated with explicit egocentric directives (e.g., standard camera-space left/right reasoning), AlloEgo-VLM maintains a stable average SR of 0.816.

To elucidate the geometric factors behind these macroscopic metrics, we provide a case-by-case analysis of Table~\ref{tab:embodied_sr}. 
Case~1 represents \textbf{Symmetrical Objects}. In layouts containing radially symmetrical entities such as tables, the absence of an intrinsic front removes spatial directionality. For example, given the query \textit{``Find the sofa right of the table,''} the baselines default to camera-space right, whereas AlloEgo-VLM may project a virtual orientation onto the object. Since both egocentric and allocentric interpretations are acceptable, all tested models achieve similarly high success rates.

Cases~2--5 represent \textbf{Asymmetrical/Directional Objects}, where AlloEgo-VLM demonstrates a clear advantage, except in Case~4. For entities with intrinsic orientation (e.g., sofas or chairs), our model dynamically infers their facing direction instead of relying solely on image-plane reasoning. While the baselines suffer from geometric confusion when objects directly face the robot, AlloEgo-VLM successfully decouples the observer viewpoint from the object-centric frame, enabling robust coordinate transformation.
The performance drop in Case 4 (Fig.~\ref{fig:scene}, left) is primarily driven by the visual ambiguity of a lateral view. When a chair faces directly sideways relative to the robot's camera, the spatial asymmetry between its left and right sides becomes visually compressed and partially occluded. This lack of distinct directional features makes it highly challenging for the model to reliably anchor the object's intrinsic coordinate axes, occasionally leading to minor frame alignment errors.

Case~5 presents a challenging human-avatar scenario with strong backlighting and the complex non-rigid structure of the human body (Fig.~\ref{fig:scene}, right). 
In the AQ experiments, all models perform poorly, while AlloEgo-VLM achieves a slightly higher SR of 0.4. 
This bottleneck suggests that tracking highly articulated human-centric reference frames under randomized camera viewpoints and varying ambient lighting remains a major challenge for embodied spatial intelligence. 
For example, even a slight change in human orientation can make it difficult to determine whether an object lies to the right or front-right of the person.

Case~6 reveals a geometric edge case related to viewpoint alignment. When the robot approaches the wheelchair from a strict rear view, the baseline models perform robustly because their egocentric axes coincidentally align with the object's global geometry, rather than through effective reference-frame transformation. Conversely, although AlloEgo-VLM incorporates explicit frame-transformation reasoning, certain viewing distances or angles can obscure the subtle visual cues that indicate the object's intrinsic orientation, resulting in a slightly lower SR.

\subsection{Additional Experiments}

Additional experiments are summarized below; see the Supplementary Material for details.

\subsubsection{Emergent Robustness Against Out-of-Distribution Inputs}
During the early training stages of AlloEgo-Adapter, we observed an intriguing vulnerability to out-of-distribution (OOD) inputs, specifically when queried about non-existing objects within an image or when prompted with pure text during simulated camera failures. Under these anomalous conditions, models trained on fewer than 2K samples exhibited rigid format imitation, blindly fabricating hallucinated descriptions to satisfy the structured spatial template without grounding. In practical robotic deployments, such unverified compliance could lead to catastrophic actions based on fabricated perceptions.

Crucially, as training progressed into Stage~3 (exceeding 2K samples), \textit{Qwen~2.5-VL-7B}~\cite{qwen2025qwen2_5_vl_7b} exhibited strong emergent capabilities. Without any explicit architectural modifications or targeted OOD supervision, the model spontaneously learned to identify missing objects and appropriately handle pure-text queries. This pivot from rote structural mimicry to genuine spatial reasoning demonstrates that robust format alignment, when supported by sufficient data scale, fosters true semantic generalization rather than mere surface-level memorization.

\subsubsection{Structured Response Design Principles}

Our structured answer format is consistently applied across AlloEgo-View to reduce view ambiguity.  
People may unconsciously adopt different reference frames due to variations in cognitive psychology and cultural background, leading to inconsistent spatial reasoning. 
Through experiments, we validate the following principles, which help ensure consistent and view-aware responses under a unified framework.

\begin{itemize}[noitemsep]
\item \textbf{Global Context:} Begin with an overall image description to establish a coherent understanding before detailed reasoning.
\item \textbf{Absolute Directions:} Provide \texttt{<Ref Abs Dir>} and \texttt{<Tgt Abs Dir>} to clarify relative directions and improve consistency in both egocentric and allocentric views.
\item \textbf{Object Positions:} Specify \texttt{<Ref Pos>} and \texttt{<Tgt Pos>} to avoid ambiguities when objects appear at distant image boundaries.
\item \textbf{Egocentric Grounding:} Use phrasing explicitly tied to the observer's perspective (e.g., ``In the image, \ldots'') to anchor reasoning.
\item \textbf{Depth Ordering:} Replace ambiguous front--back terms with explicit relational statements (e.g., ``between the observer and \ldots'') to reduce 3D ambiguity.
\item \textbf{Allocentric Grounding:} Define allocentric relations with respect to the reference object's intrinsic orientation when available; otherwise, avoid assigning object-centered directions to non-directional objects.
\item \textbf{Simplified Labeling:} Avoid labels such as ``looking back,'' which may vary across cultural or individual interpretations.
\item \textbf{Non-directional Objects:} Use objects without intrinsic orientation (e.g., balls, plants, fruits) for egocentric relations only, and avoid treating them as allocentric reference objects.
\item \textbf{Geographic Directions:} Exclude absolute geographic references (e.g., north, south, east, west) as they lie outside the scope of this work.
\end{itemize}


\section{Conclusions}

In this work, we address spatial semantic ambiguity in VLMs by distinguishing allocentric and egocentric reference frames. 
We introduce a structured spatial representation framework, the AlloEgo-View dataset, and a multi-stage QLoRA-based fine-tuning pipeline, collectively enabling AlloEgo-VLM with strong spatial disambiguation capability.

Experimental results show that AlloEgo-VLM substantially outperforms state-of-the-art models, including \textit{GPT-4o}, \textit{Llama~3.2-Vision}, and \textit{Gemma~3-Vision}, on spatial reasoning tasks. 
Notably, \textit{Qwen~2.5-VL-7B} trained on sufficiently large datasets exhibits emergent capabilities, such as detecting missing objects and correctly answering pure text queries, while maintaining strong performance across diverse query types.
Our findings highlight the importance of dataset scale and structured supervision for robust spatial reasoning. 
Unlike memory-intensive \textbf{MoE} approaches~\cite{jiang_mixtral_2024}, our lightweight classifier introduces minimal overhead, enabling more reliable deployment in embodied AI and robotics.

\bibliography{egbib}

\appendix
\section{Appendix}

\subsection{Prompt for RGB Image Selection}
\label{app:prompt_for_image_filtering}

The prompt for RGB image selection in Sec.~\ref{sec:rgb-select} is shown below:
\vspace{-6pt}

\noindent
\rule{\columnwidth}{0.4pt}
\noindent
\textbf{Prompt:}
Please determine whether the image meets the following criteria:
\begin{enumerate}[noitemsep]
    \item The image contains approximately 2 to 6 distinct, identifiable objects or entities.
    \item The background is relatively clean and uncluttered.
    \item The scene could potentially lead to referential ambiguity in natural language descriptions due to varied perspectives or viewpoints.
\end{enumerate}
These images will be used to generate question-answer pairs related to referential ambiguity. Respond only with ``Yes'' or ``No.''

\vspace{-0.2cm}
\noindent
\rule{\columnwidth}{0.4pt}

\subsection{Prompt for Structured Spatial Answer Generation}
\label{app:assistant_prompt}

To generate the initial structured spatial answer (Step 1) in Sec. \ref{sec:answer-gen}, we employ GPT-4o with the following prompt:

\vspace{-10pt}

\medskip
\noindent
\rule{\columnwidth}{0.4pt}
\noindent
\textbf{Prompt:}
Please follow the instructions below to describe the direction and spatial relationship between two objects:

\begin{enumerate}[noitemsep]
    \item \textbf{Overall Image Description:} Based on the visual content of the image, provide a comprehensive description of the entire scene, including:
    \begin{itemize}[noitemsep]
        \item The overall setting and background context (e.g., indoor/outdoor, urban/natural environment).
        \item The number and types of visible objects (e.g., people, vehicles, buildings, furniture).
        \item The spatial distribution and relative positions of objects within the scene (e.g., clustered on the left, evenly spread across the image).
        \item Any prominent visual structures or compositional features that help define the spatial layout (e.g., roads, walls, floor lines, depth cues in the background).
    \end{itemize}

    \item \textbf{Reference Object Absolute Orientation:}
    \begin{itemize}[noitemsep]
        \item Do not compare it to any other object.
        \item Only describe its own directional properties.
        \item Use absolute terms such as: ``facing left'', ``facing right'', ``facing upward'', ``facing downward'', ``facing the observer'', ``facing away from the observer'', or ``this object has no inherent direction''.
        \item An object is considered to have inherent directionality if its front, back, left, or right side can be visually inferred from its shape, posture, or design (e.g., a person, car, or animal). Objects like chairs or cups may have direction depending on their orientation. If no such direction is visually evident, state ``this object has no inherent direction''.
    \end{itemize}

    \item \textbf{Target Object Absolute Orientation:}
    \begin{itemize}[noitemsep]
        \item Same rules as the Reference Object: do not compare to any other object and only describe its own directionality.
    \end{itemize}

    \item \textbf{Relative Position of the Target Object with Respect to the Reference Object:}
    \begin{itemize}[noitemsep]
        \item If at least one of the two objects has directionality, provide both \textbf{Egocentric} and \textbf{Allocentric} perspectives.
        \item If neither of the two objects has directionality, provide \textbf{Egocentric} only.
        \item \textbf{Egocentric Description (Observer-Centered):}
        \begin{enumerate}
            \item Treat both the Reference Object and the Target Object as 2D bounding boxes in screen space.
            \item Think of the image as a nine-square grid. Describe their \textbf{individual screen positions} using: ``upper left'', ``upper center'', ``upper right'', ``center left'', ``center'', ``center right'', ``lower left'', ``lower center'', ``lower right''.
            \item Describe their \textbf{relative screen positions} using: 
            \begin{quote}
            \raggedright
            ``\texttt{<Target Object>} is at the same position as the \texttt{<Reference Object>}'', ``\texttt{<Target Object>} is above the \texttt{<Reference Object>}'', ``\texttt{<Target Object>} is below the \texttt{<Reference Object>}'', ``\texttt{<Target Object>} is to the left of the \texttt{<Reference Object>}'', ``\texttt{<Target Object>} is to the right of the \texttt{<Reference Object>}'', ``\texttt{<Target Object>} is to the upper left of the \texttt{<Reference Object>}'', ``\texttt{<Target Object>} is to the lower left of the \texttt{<Reference Object>}'', ``\texttt{<Target Object>} is to the upper right of the \texttt{<Reference Object>}'', ``\texttt{<Target Object>} is to the lower right of the \texttt{<Reference Object>}''.
            \end{quote}
            \item If a depth relationship is visible (occlusion or perspective cues), indicate: 
            \begin{quote}
            ``\texttt{<Target Object>} is between the observer and the \texttt{<Reference Object>}'', ``\texttt{<Reference Object>} is closer to the observer than the \texttt{<Target Object>}'', or ``\texttt{<Target Object>} is farther from the observer than the \texttt{<Reference Object>}''.
            \end{quote}
        \end{enumerate}

        \item \textbf{Allocentric Description (Reference Object-Centered):}
        \begin{enumerate}[noitemsep]
            \item Use the orientation of the Reference Object to describe the location of the Target Object.
            \item Use spatial terms such as: in front of, behind, to the left of, to the right of, diagonally front-left, etc.
        \end{enumerate}
    \end{itemize}

    \item \textbf{Response Format:}
    \begin{itemize}[noitemsep]
        \raggedright
        \item 
        Overall Image Description: \texttt{<Overall Image Description>}
        \item 
        Reference Object: \texttt{<Reference Object>}
        \item 
        Target Object: \texttt{<Target Object>}
        \item
        Reference Object Absolute Direction: \texttt{<Reference Object>} is facing \texttt{<Direction>}
        \item
        Target Object Absolute Direction: \texttt{<Target Object>} is facing \texttt{<Direction>}
        \item 
        Perspective: Egocentric (from the observer's point of view)
        \item
        Answer: The \texttt{<Reference Object>} is on the \texttt{<Position>} of the image, the \texttt{<Target Object>} is on the \texttt{<Position>} of the image, and in the image, \texttt{<Target Object>} is \texttt{<Direction>} \texttt{<Reference Object>}.
        \item 
        Perspective: Allocentric (from the \texttt{<Reference Object>}'s point of view)
        \item
        Answer: From the \texttt{<Reference Object>}'s perspective, the \texttt{<Target Object>} is \texttt{<Direction>} of the \texttt{<Reference Object>}.
    \end{itemize}
\end{enumerate}

\medskip
\noindent
\textbf{Example 1: }  

\emph{Overall Image Description}: The image shows a heartwarming scene of a golden retriever lying on the floor next to a gray and white cat. The cat is gently nuzzling the dog's face, creating a sense of affection between the two pets. In front of them is a white bowl, likely containing food, suggesting they might be sharing a meal. The setting appears to be a cozy indoor space with white cabinets, shelves with books or papers, and a light-colored floor.

\emph{Reference Object}: Dog

\emph{Target Object}: Cat

\emph{Reference Object Absolute Direction}: The dog is facing the observer

\emph{Target Object Absolute Direction}: The cat is facing left

\emph{Perspective}: Egocentric (from the observer's point of view)

\emph{Answer}: The dog is on the left middle of the image, the cat is on the right middle of the image, and in the image, cat is to the right of dog.

\emph{Perspective}: Allocentric (from the Dog's point of view)

\emph{Answer}: From the dog's perspective, the cat is on its right side.

\medskip
\noindent
\textbf{Example 2: }  

\emph{Overall Image Description}: This image shows two plastic buckets placed on a smooth, light-colored surface. The bucket in the foreground is blue with a white handle, while the bucket behind it is red, also with a white handle. The blue bucket is positioned slightly to the left and in front of the red one, creating a clear sense of depth and perspective. The scene appears to be well-lit, likely photographed indoors or in a shaded outdoor setting. The overall composition is simple and minimalistic. 

\emph{Reference Object}: Red bucket

\emph{Target Object}: Blue bucket

\emph{Reference Object Absolute Direction}: The red bucket has no inherent direction

\emph{Target Object Absolute Direction}: The blue bucket has no inherent direction

\emph{Perspective}: Egocentric (from the observer's point of view)

\emph{Answer}: The red bucket is on the upper right of the image, the blue bucket is on the center left of the image, and in the image, the red bucket is to the upper right of the blue bucket. 

\emph{Perspective}: Allocentric (from the red bucket's point of view)

\emph{Answer}: No Allocentric.

\vspace{-0.2cm}
\noindent
\rule{\columnwidth}{0.4pt}

\vspace{4pt}
\subsection{Examples of Spatial Answers}
\label{app:example_TF}

In the following, we present the preliminary answer generated by GPT-4o for the example shown in Fig.~\ref{fig:3}.

\vspace{-10pt}

\medskip
\noindent
\rule{\columnwidth}{0.4pt}

\noindent\emph{Overall Image Description}: The image shows an indoor setting with two people. A man in a boxing stance is on the left, and a woman holding a wine glass is on the right. They are standing in front of a door, with some office equipment visible on the left side of the image. The room appears to be well-lit, with a neutral-colored wall and floor.

\noindent\emph{Reference Object}: 
\textcolor{blue}{The man in a boxing stance}

\noindent\emph{Target Object}: 
\textcolor{green}{The woman holding a wine glass}

\noindent\emph{Reference Object Absolute Direction}: 
The man in a boxing stance is facing the observer

\noindent\emph{Target Object Absolute Direction}: 
The woman holding a wine glass is facing the observer

\noindent\emph{Perspective}: 
Egocentric (from the observer's point of view)

\noindent\emph{Answer}: 
The man in a boxing stance is on the \colorbox{orange}{left} of the image, the woman holding a wine glass is on the \colorbox{orange}{right} of the image, and in the image, the woman holding a wine glass is to the right of the man in a boxing stance.

\noindent\emph{Perspective}: 
Allocentric (from the man in a boxing stance's point of view)

\noindent\emph{Answer}: 
From the man in a boxing stance's perspective, the woman holding a wine glass is to his \colorbox{cyan}{right.}

\vspace{-0.4cm}
\noindent
\rule{\columnwidth}{0.4pt}

\medskip
The preliminary answer above is then refined through the steps described in Sec.~\ref{sec:answer-gen}. 
Specifically, the orange-highlighted parts are corrected in Step~2, the yellow-highlighted parts are added in Step~3, and the blue-highlighted parts are manually corrected in Step~4:

\vspace{-10mm}

\medskip
\noindent
\rule{\columnwidth}{0.4pt}

\noindent\emph{Overall Image Description}: 
The image shows an indoor setting with two people. A man in a boxing stance is on the left, and a woman holding a wine glass is on the right. They are standing in front of a door, with some office equipment visible on the left side of the image. The room appears to be well-lit, with a neutral-colored wall and floor.

\noindent\emph{Reference Object}: 
\textcolor{blue}{The man in a boxing stance}

\noindent\emph{Target Object}: 
\textcolor{green}{The woman holding a wine glass}

\noindent\emph{Reference Object Absolute Direction}: 
The man in a boxing stance is facing the observer

\noindent\emph{Target Object Absolute Direction}: 
The woman holding a wine glass is facing the observer

\noindent\emph{Perspective}: 
Egocentric (from the observer's point of view)

\noindent\emph{Answer}: 
The man in a boxing stance is on the \colorbox{orange}{center left} of the image, the woman holding a wine glass is on the \colorbox{orange}{center right} of the image, and in the image, the woman holding a wine glass is to the right of the man in a boxing stance. 
\hl{The man in a boxing stance is between the observer and the woman holding a wine glass.}

\noindent\emph{Perspective}: 
Allocentric (from the man in a boxing stance's point of view)

\noindent\emph{Answer}: 
From the man in a boxing stance's perspective, the woman holding a wine glass is to his \colorbox{cyan}{back left.}

\vspace{-0.3cm}
\noindent
\rule{\columnwidth}{0.4pt}

\subsection{PosDir Algorithm}\label{app:Spatial_Relation_Algorithm}
\begin{algorithm}[H]
\caption{Grid-Based Spatial Relation Extraction}
\label{alg:spatial-relation}
\begin{algorithmic}[1]
\Require Reference object bounding box $BBOX_r = (x_{r1}, y_{r1}, x_{r2}, y_{r2})$, Target object bounding box $BBOX_t = (x_{t1}, y_{t1}, x_{t2}, y_{t2})$, Grid indices $grids \subseteq \{1,\dots,9\}$
\Ensure Description of the reference object $d_r$, Description of the target object $d_t$, Relative relation $d_{rel}$

\State Determine the grid index set $g_r \subseteq grids$ covered by the rectangle $BBOX_r = (x_{r1}, y_{r1}, x_{r2}, y_{r2})$
\State Determine the grid index set $g_t \subseteq grids$ covered by the rectangle $BBOX_t = (x_{t1}, y_{t1}, x_{t2}, y_{t2})$

\If{$g_r = \emptyset$ or $g_t = \emptyset$}
    \State $d_r \gets$ ``Unknown'', $d_t \gets$ ``Unknown''
\Else
    \State Look up the corresponding description for $g_r$ in Table~\ref{tab:grid-mapping}, assign to $d_r$
    \State Look up the corresponding description for $g_t$ in Table~\ref{tab:grid-mapping}, assign to $d_t$
\EndIf
\State Compute the center of the reference object bounding box $(x_r, y_r) \gets \left(\frac{x_{r1} + x_{r2}}{2}, \frac{y_{r1} + y_{r2}}{2}\right)$
\State Compute the center of the target object bounding box $(x_t, y_t) \gets \left(\frac{x_{t1} + x_{t2}}{2}, \frac{y_{t1} + y_{t2}}{2}\right)$
\State $\Delta x \gets x_t - x_r$
\If{$|\Delta x| > \gamma$}
    \If{$\Delta x > 0$}
        \State $d_{rel} \gets$ ``Right''
    \Else
        \State $d_{rel} \gets$ ``Left''
    \EndIf
\EndIf
\State \Return $(d_r, d_t, d_{rel})$
\end{algorithmic}
\end{algorithm}

\vspace{-4mm}

\begin{table}[H]
\footnotesize
\centering
\begin{tabular}{ll}
\toprule
Grid set $grids$ & Description $d$ \\
\midrule
$[2]$ / $[4]$ / $[5]$ / $[6]$ / $[8]$& Top / Left / Center / Right / Bottom\\
$[1]$ / $[3]$ & Top-left / Top-right\\
$[7]$ / $[9]$ & Bottom-left / Bottom-right\\
\midrule
$[1,2]$, $[1,4]$ / $[2,3]$, $[3,6]$ & Top-left / Top-right\\
$[4,7]$, $[7,8]$ / $[6,9]$, $[8,9]$ & Bottom-left / Bottom-right\\
$[2,5]$ / $[5,8]$ & Upper-center / Lower-center\\
$[4,5]$ / $[5,6]$& Left-center / Right-center\\
\midrule
$[1,2,3]$ / $[7,8,9]$ & Top / Bottom\\
$[4,5,6]$, $[2,5,8]$ & Center \\
$[1,4,7]$ / $[3,6,9]$ & Left /Right\\
\midrule
$[1,2,4,5]$ / $[2,3,5,6]$& Upper-left-center / Upper-right-center\\
$[4,5,7,8]$ / $[5,6,8,9]$& Lower-left-center / Lower-right-center\\
\midrule
$[1,2,3,4,5,6]$ / $[4,5,6,7,8,9]$& Upper-center / Lower-center\\
$[1,2,4,5,7,8]$ / $[2,3,5,6,8,9]$& Left-center /Right-center\\
\midrule
$[1,2,3,4,5,6,7,8,9]$ & Center \\
\bottomrule
\end{tabular}
\vspace{8pt}
\captionsetup{labelfont={color=mycaptionblue,bf}, textfont={color=mycaptionblue}}
\caption{Mapping rules between grid sets and spatial descriptions}
\label{tab:grid-mapping}
\end{table}

\subsection{DepOrd Algorithm}\label{app:Depth_Ordering_Algorithm}
\vspace{-10pt}
\begin{algorithm}[H]
\caption{Depth-Based Relative Position Description}
\label{alg:depth-relative-position}
\begin{algorithmic}[1]
\Require Depth values of Reference object $D_r = \{d_{r1}, d_{r2}, \dots, d_{rn}\}$, Depth values of Target object $D_t = \{d_{t1}, d_{t2}, \dots, d_{tm}\}$, Threshold $\gamma$
\Ensure Description of spatial depth ordering $desc$

\State Compute the average depth of the Reference object: 
\[
\bar{D}_r \gets \frac{1}{n} \sum_{i=1}^{n} d_{ri}
\]

\State Compute the average depth of the Target object: 
\[
\bar{D}_t \gets \frac{1}{m} \sum_{i=1}^{m} d_{ti}
\]

\State Compute depth difference: $\Delta D \gets \bar{D}_r - \bar{D}_t$

\If{$\Delta D > \gamma$}
    \State $desc \gets$ ``\texttt{<Target Object> is between the observer and the <Reference Object>.}''
    \State Optionally add:
    \State ``\texttt{<Reference Object> is farther to the observer than the <Target Object>.}''
    \State ``\texttt{<Target Object> is closer from the observer than the <Reference Object>.}''
\ElsIf{$\Delta D < -\gamma$}
    \State $desc \gets$ ``\texttt{<Reference Object> is between the observer and the <Target Object>.}''
    \State Optionally add:
    \State ``\texttt{<Target Object> is farther to the observer than the <Reference Object>.}''
    \State ``\texttt{<Reference Object> is closer from the observer than the <Target Object>.}''
\EndIf

\State \Return $desc$

\end{algorithmic}
\end{algorithm}


\subsection{Spatial Query Template List}
\label{app:user_input_collection}

In Sec.~\ref{sec:question-gen}, we use 17 predefined templates to generate spatial queries that are consistent yet diverse. 
These templates are listed below:
\begin{enumerate}[noitemsep]
    \raggedright
    \item Where is the \texttt{<Target Object>} in relation to the \texttt{<Reference Object>}?
    \item How are the \texttt{<Target Object>} and \texttt{<Reference Object>} positioned?
    \item Can you describe where the \texttt{<Target Object>} is?
    \item How would you describe the position of the \texttt{<Target Object>} compared to the \texttt{<Reference Object>}?
    \item What is the location of the \texttt{<Target Object>} relative to the \texttt{<Reference Object>}?
    \item Where would you say the \texttt{<Target Object>} is placed?
    \item Tell me how the \texttt{<Target Object>} and the \texttt{<Reference Object>} are arranged.
    \item If someone asked you where the \texttt{<Target Object>} is, what would you say?
    \item Where is the \texttt{<Target Object>} located with respect to the \texttt{<Reference Object>}?
    \item What is the spatial relationship between the \texttt{<Target Object>} and the \texttt{<Reference Object>}?
    \item Can you point out where the \texttt{<Target Object>} is compared to the \texttt{<Reference Object>}?
    \item Where do you see the \texttt{<Target Object>}?
    \item What is the position of the \texttt{<Target Object>} in relation to the other object?
    \item Where does the \texttt{<Target Object>} appear to be?
    \item Which side of the \texttt{<Reference Object>} is the \texttt{<Target Object>} on?
    \item How would you explain where the \texttt{<Target Object>} is to someone else?
    \item Looking at the scene, where is the \texttt{<Target Object>}?
\end{enumerate}


\subsection{Prompt for View Ambiguity Rate Test}
\label{app:Prompt_for_judging_view_awareness}

In Table~\ref{tab:Ambiguity_rate}, the View Ambiguity Rate (VAR) is evaluated using GPT-4o with the following prompt.

\vspace{-10pt}

\medskip
\noindent
\rule{\columnwidth}{0.4pt}
\noindent
\textbf{Prompt:}
Determine whether the following response contains a Reference frame.
Rule: If the response includes the phrase ``from the perspective of ...'' (or similar wording), answer ``yes''. Otherwise, answer ``no''.  
Only output ``yes'' or ``no''---no extra text.

\vspace{-0.2cm}
\noindent
\rule{\columnwidth}{0.4pt}



\subsection{Prompt for Automated Model Scoring}
\label{app:score_prompt}

In Table~\ref{tab:model_comparison}, the four metrics 
{AD\_RO}, {AD\_TO}, {Ego}, and {Allo} 
are evaluated using GPT-4o with the following prompt:

\vspace{-10pt}

\medskip
\noindent
\rule{\columnwidth}{0.4pt}
\noindent
\textbf{Prompt:}

You are a semantic evaluation expert.

\medskip

Assistant Response: \{AR["answers"][i]\} 

Ground Truth Answer: \{GT["answers"][i]\}

\noindent
Evaluate their similarity in four specific aspects:

\begin{enumerate}[noitemsep]
    \item \textbf{Reference Object Absolute Direction Score: X / 10}
    \begin{itemize}
        \item Assess how accurately the Reference Object Absolute Direction in the Assistant Response matches the Ground Truth Answer.
        \item Consider semantic correctness, direction consistency, and clarity.
    \end{itemize}

    \item \textbf{Target Object Absolute Direction Score: X / 10}
    \begin{itemize}
        \item Assess how accurately the Target Object Absolute Direction in the Assistant Response matches the Ground Truth Answer.
        \item Consider semantic correctness, direction consistency, and clarity.
    \end{itemize}

    \item \textbf{Egocentric Answer Score: X / 10}
    \begin{itemize}
        \item Assess how accurately the Egocentric answer in the Assistant Response matches the Ground Truth Answer.
        \item Consider whether the spatial relationship and positional details match.
    \end{itemize}

    \item \textbf{Allocentric Answer Score: X / 10}
    \begin{itemize}
        \item Assess how accurately the Allocentric answer in the Assistant Response matches the Ground Truth Answer.
        \item Consider whether the description from the reference object's perspective is semantically correct and consistent.
    \end{itemize}
\end{enumerate}

\noindent
For each category, score from 1 to 10 (10 = completely accurate and aligned; 1 = entirely incorrect).

\noindent
Then, for each category, provide a clear explanation of your reasoning, addressing:
\begin{itemize}[noitemsep]
    \item Whether meanings align
    \item Whether any important details are missing or incorrect
    \item Whether the response is misleading
    \item Clarity of expression
\end{itemize}

\noindent
\textbf{Output Format:}\\
Reference Object Absolute Direction Score: X / 10\\
Explanation: ...\\
Target Object Absolute Direction Score: X / 10\\
Explanation: ...\\
Egocentric Answer Score: X / 10\\
Explanation: ...\\
Allocentric Answer Score: X / 10\\
Explanation: ...

\vspace{-0.2cm}
\noindent
\rule{\columnwidth}{0.4pt}


\subsection{Experiments with Various Inputs}
\label{app:Interference_issues}

In these experiments, we feed various queries to our model and make observations.

\begin{figure}[H]
    \centering
    \begin{tabular}{c}
        \bmvaHangBox{\includegraphics[width=0.8\textwidth]{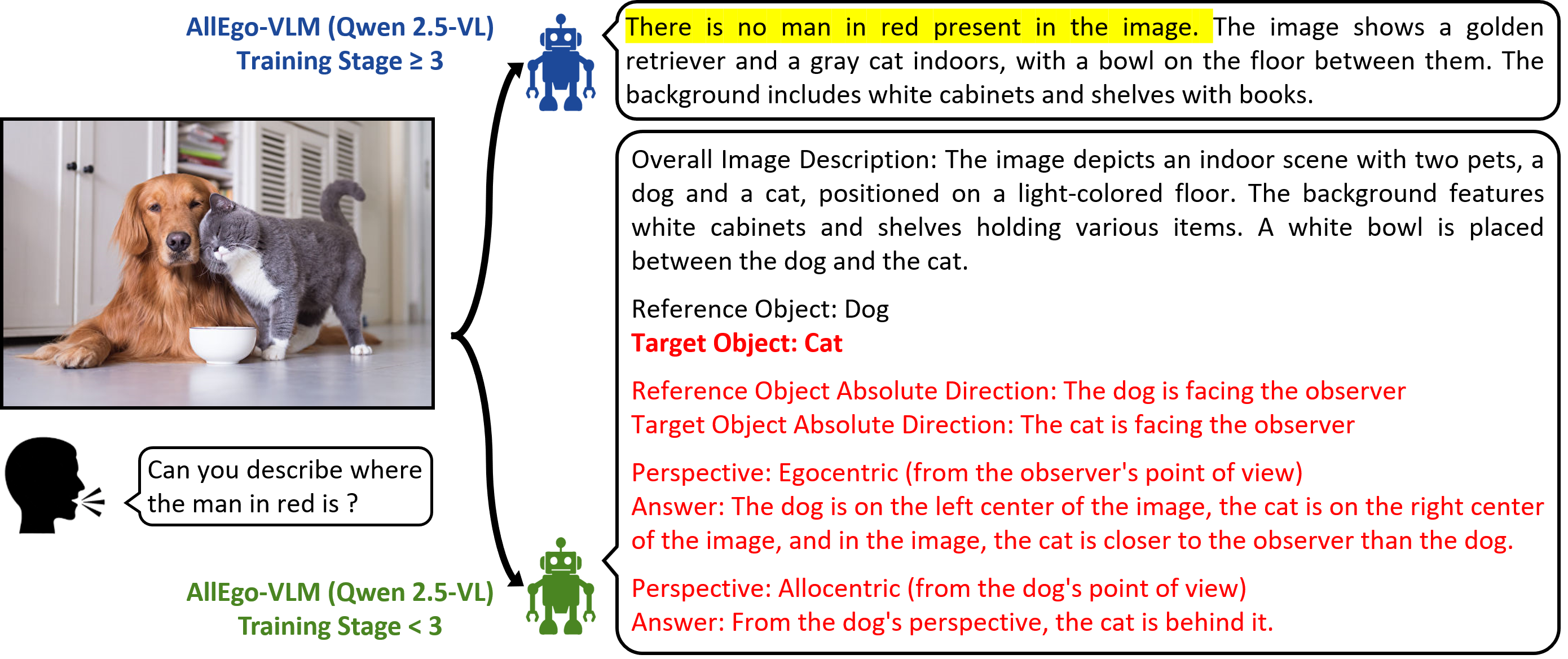}} \\
        \parbox{0.8\textwidth}{\centering \textcolor{mycaptionblue}{\refstepcounter{subfigure}\label{app:FD10D1a}\textbf{(a)} The queried object is not present in the input image.}} \\[15pt]
        
        \bmvaHangBox{\includegraphics[width=0.8\textwidth]{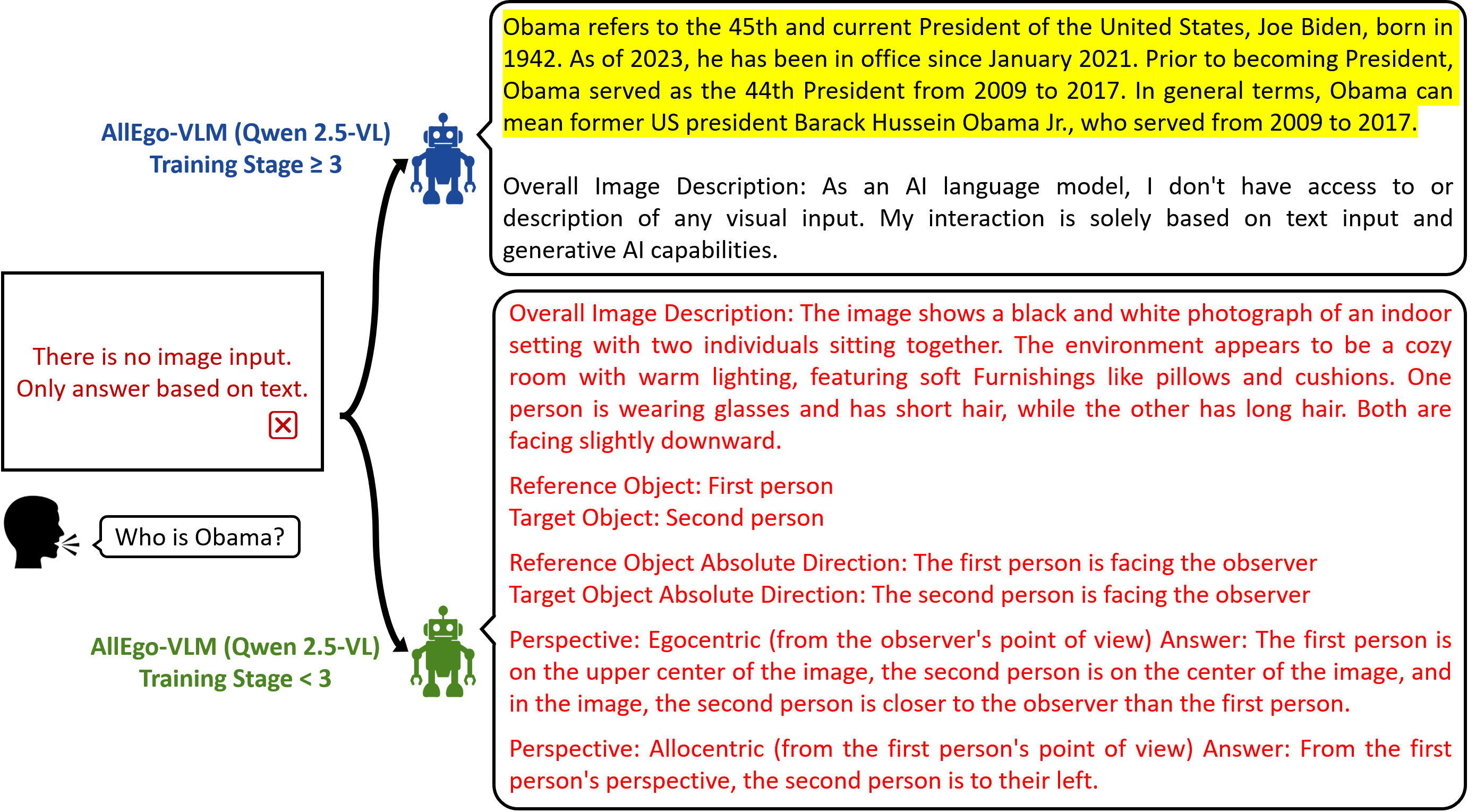}} \\
        \parbox{0.8\textwidth}{\centering \textcolor{mycaptionblue}{\refstepcounter{subfigure}\label{app:FD10D1b}\textbf{(b)} No image input.}}
    \end{tabular}
    \vspace{10pt}
    \captionsetup{labelfont={color=mycaptionblue,bf}, textfont={color=mycaptionblue}}
    \caption{Examples of emergent reasoning in AlloEgo-VLM (with Qwen2.5-VL as the backbone). Behaviors after and before training stage~3 are shown.}
    \label{fig:FD10D}
\end{figure}

\begin{itemize}
\item 
When given an input image and a question about an object absent from the image, AlloEgo-VLM (with Qwen2.5-VL as the backbone) trained on a sufficiently large dataset (after training stage 3 in Fig.~\ref{fig:multi_stage}) correctly identifies the object's absence (highlighted in yellow in Fig.~\ref{app:FD10D1a}). In contrast, the model trained on an insufficient dataset (before stage 3) responds solely based on the visible content, disregarding the query (shown in red in Fig.~\ref{app:FD10D1a}).
\item 
When given only textual input without an accompanying image, the model trained on a sufficiently large dataset accurately answers the question and explicitly acknowledges the lack of visual input (highlighted in yellow in Fig.~\ref{app:FD10D1b}). In contrast, the model trained on an insufficient dataset hallucinates, generating fabricated content unrelated to the question (shown in red in Fig.~\ref{app:FD10D1b}), revealing its tendency to produce ungrounded responses without visual context.
\end{itemize}

\subsection{Structured Answer Format Design Principle}
\label{app:Answer format design techniques}

Here, we explain our design principle of the structured answer format.

\begin{enumerate}
\item 
\textbf{Global Context (Overall Image Description):} Inspired by \textit{Chain-of-Thought} \cite{Ji2025CoTSpatial} reasoning (e.g., having a language model write out detailed calculations improves accuracy on math tasks), we require the model to provide an overall description of the image at the beginning. This helps the model understand the global structure and context before proceeding to detailed reasoning.
\item 
\textbf{Absolute Directions \texttt{<Ref Abs Dir>} and \texttt{<Tgt Abs Dir>}:} Providing the absolute directions of the reference and target objects not only helps models better under- stand their spatial relationships in the image, but also improves the accuracy of subsequent egocentric and allocentric view descriptions. These absolute directions are crucial for correctly inferring relative directions.
\item 
\textbf{Object Positions (\texttt{<Ref Pos>} and \texttt{<Tgt Pos>}):}  
Specifying object positions is crucial when the two objects are far apart in the image.  
For example, if a man is at the left boundary and a woman at the right boundary, explicit position information is essential; otherwise, describing their relationship \texttt{<Ego Rel Dir>} becomes practically infeasible.  
Including positions is therefore indispensable to avoid such ambiguities.
\item 
\textbf{Egocentric Grounding:} 
Even though the egocentric view is initially prompted with {``from the observer's point of view,''} it is necessary to explicitly instruct the model to phrase answers like {``In the image, ...''} when resolving \texttt{<Ego Rel Dir>}. This step is essential to ensure that all reasoning is firmly anchored in the observer's visual perspective.
\item 
\textbf{Depth Ordering:} 
If Step 3 (Depth Enhancement) is not implemented, ambiguities will arise. In an image, vertical relations (up–down) pose no ambiguity due to physical constraints. However, the remaining relations—front–back and left–right—are prone to ambiguity. Step 2 already resolves the 2D left–right issue, while Step 3 specifically addresses the 3D depth (front–back) relation to avoid such problems. Here, we adopt expressions such as ``The \texttt{<Ref Obj>} is between the observer and the \texttt{<Tgt Obj>}'' or ``The \texttt{<Tgt Obj>} is farther from the observer than the \texttt{<Ref Obj>},'' instead of simply saying ``in front'' or ``behind.'' 
For example, if the red bucket is in front of the blue bucket, the ambiguity arises because, even from an egocentric viewpoint, cultural backgrounds may differ: some people consider objects closer to the observer as being ``in front,'' while others regard them as ``behind.''
Another example is: ``The man is facing the observer, the woman is facing away from the observer, and the man is in front of the woman.''
In this case, it is also unclear whether the man is actually closer to the observer or the woman is closer, leading to ambiguity.
\item 
\textbf{Allocentric Challenges:} 
There is currently no effective solution for allocentric view descriptions, and no model can accurately and reliably label them. Many objects cannot have their orientation directly defined from RGBD images—for example, a sofa, a monitor, or a microwave. Even if these objects inherently lack a clear orientation, humans tend to infer their direction based on usage context. Existing VLMs perform poorly in this regard because they have not been trained with knowledge of ``contextually defined orientation'' during usage. This also highlights a key challenge addressed in this paper: not all cases can rely on tools to label data for improving VLM capabilities, which further validates the significance of our contribution.    
\item 
\textbf{Simplifying Labeling:} 
During the labeling process, we tried to avoid overly complex scenarios—for example, when a person is ``looking back.'' In such cases, it is difficult to define whether the person is facing toward or away from the observer, as interpretations can vary across individuals and cultures. Therefore, we excluded images of this type to keep the absolute direction as simple and consistent as possible. In the process of creating this dataset, similar issues may arise frequently. While some tasks may have clear definitions in spatial studies or cognitive psychology, here we rely solely on humans' intuitive understanding of direction. This remains a problem worth addressing in the future, with the potential for clearer definitions.
    
    \item \textbf{Non-directional Objects:} Some objects (e.g., soccer balls, potted plants, stones, fruits) have no intrinsic orientation and are excluded from direction annotation.
    
    \item \textbf{Geographic Directions:} Absolute geographic directions (e.g., north, south, east, west) are beyond the scope of this work.
    
\end{enumerate}
\include{appendix}
\end{document}